\pdfoutput=1
\documentclass[11pt]{article}

\usepackage{acl}

\usepackage{times}
\usepackage{latexsym}

\usepackage[T1]{fontenc}
\usepackage[utf8]{inputenc}

\usepackage{microtype}

\usepackage{times}
\usepackage{soul}
\usepackage{url}
\usepackage{subcaption}
\usepackage{amsmath, amsfonts}
\usepackage{enumitem}
\usepackage{geometry}
\usepackage{pdflscape}
\usepackage{adjustbox}
\usepackage{multirow}
\usepackage{philex}
\usepackage{caption}
\usepackage{booktabs}
\usepackage{makecell}
\usepackage{algorithm}
\usepackage{algorithmic}
\usepackage{amsfonts}
\usepackage[switch]{lineno}
\usepackage{pgfplots}
\usepackage{graphicx}
\usepackage{atbegshi,picture}
\usepackage{fancyhdr}
\title{TaxoLLaMA: WordNet-based Model \\for Solving Multiple Lexical Semantic Tasks}

\author{\textbf{Viktor Moskvoretskii$^{1,2}$, 
Ekaterina Neminova$^1$, Alina Lobanova$^{1}$,} \\ \textbf{Alexander Panchenko$^{2,3}$, and Irina Nikishina$^{4}$}\\
$^1$HSE University,
$^2$Skoltech,
$^3$AIRI,
$^4$Universit{\"a}t Hamburg \\
\href{mailto:v.moskvoretskii@skol.tech}{\{v.moskvoretskii, a.panchenko\}@skol.tech}, \hspace{3pt} \href{mailto:esneminova@edu.hse.ru}
{\{esneminova, alobanova\}@edu.hse.ru}, \\ \hspace{3pt} \href{mailto:irina.nikishina@uni-hamburg.de}{irina.nikishina@uni-hamburg.de}}

\begin{document}
\maketitle
\begin{abstract}
\let\thefootnote\relax\footnote{Accepted to ACL Main 2024}
In this paper, we explore the capabilities of LLMs in capturing lexical-semantic knowledge from WordNet on the example of the \mbox{LLaMA-2-7b} model and test it on multiple lexical semantic tasks. As the outcome of our experiments, we present \text{TaxoLLaMA}, the ``all-in-one'' model for taxonomy-related tasks, lightweight due to 4-bit quantization and LoRA. \text{TaxoLLaMA} achieves 11 SOTA results, and 4 top-2 results out of 16 tasks on the Taxonomy Enrichment, Hypernym Discovery, Taxonomy Construction, and Lexical Entailment tasks. Moreover, it demonstrates a very strong zero-shot performance on Lexical Entailment and Taxonomy Construction with no fine-tuning. We also explore its hidden multilingual and domain adaptation capabilities with a little tuning or few-shot learning. All datasets, code, and pre-trained models are available online.\footnote{\url{https://github.com/VityaVitalich/TaxoLLaMA}}
\end{abstract}

\rhead{ACL Main 2024}
\section{Introduction}

Recent studies in Natural Language Processing widely utilize Large Language Models (LLMs) for their capability to store extensive knowledge \citep{sun2023head, kauf2023event, tang2023llamas} and to adapt quickly to different tasks via in-context learning without backpropagation \citep{dong2023survey}.
However, the application of LLMs to the classical lexical semantic tasks still remains understudied: for instance, no recent experiments with LLMs have been performed for the Hypernym Discovery task \cite{camacho-collados-etal-2018-semeval} for different domains and languages. In Taxonomy Enrichment, LLMs are mostly used to extract vector representations which are further processed with a complex pipeline \cite{jiang2022taxoenrich}.

%Delving into these tasks could yield improved performance and challenge the models with authentic linguistic problems. Furthermore, this suggests a valuable direction for further investigation of LLMs within the realm of traditional linguistic research.

\begin{figure}[h]
    \centering
\includegraphics[width=0.3\textwidth]{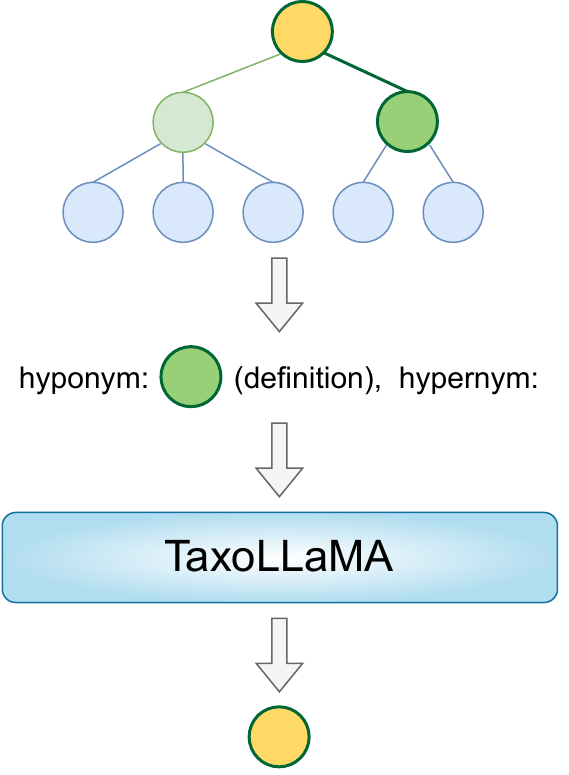}
    \caption{Training procedure of \text{TaxoLLaMA}: hypernym relations from the WordNet are linearized and fed into an LLM model. The model aims at generating the correct hypernym(s) as output.}
    \label{fig:taxllama}
    \vspace{-0.5cm}
\end{figure}

Our work aims to investigate the capabilities of LLMs in addressing four tasks requiring taxonomic knowledge: Hypernym Discovery, Taxonomy Enrichment, Lexical Entailment, and Taxonomy Construction. 
We hypothesize that the model finetuned with hypernym (IS-A relationships) would be useful for solving taxonomy-related tasks.
To verify this hypothesis, we develop a method inspired by \cite{moskvoretskii-etal-2024-large-language} to compile a taxonomy-focused instruction tuning dataset, sourced from  English WordNet \citep{miller1998wordnet}, to bring the implicit word knowledge of an LLM to the forefront when addressing lexical semantic tasks.

Having trained our model in this specialized setting, we are releasing the \text{TaxoLLaMA} --- the finetuned version of the LLaMA-2-7b model \citep{llama2} --- that is capable of solving tasks requiring taxonomic knowledge. Figure \ref{fig:taxllama} presents the main idea of the model finetuning process. \text{TaxoLLaMA} operates effectively in a zero-shot setting, surpassing SOTA results in Lexical Entailment and Taxonomy Construction. With additional tuning, it also achieves SOTA performance in the Hypernym Discovery task across several languages and in half of the Taxonomy Enrichment tasks. Furthermore, we have optimized \text{TaxoLLaMA} to be lightweight through 4-bit quantization \citep{dettmers2023qlora} and the application of LoRA \citep{hu2021lora}, making it feasible to run on GPU devices with only 4.8Gb of GPU for forward pass and 5.5Gb for fine-tuning, ensuring its accessibility for widespread use, e.g. using Colab\footnote{\url{https://colab.research.google.com}}.

\begin{figure*}[t]
    \centering
\includegraphics[width=0.99\textwidth]{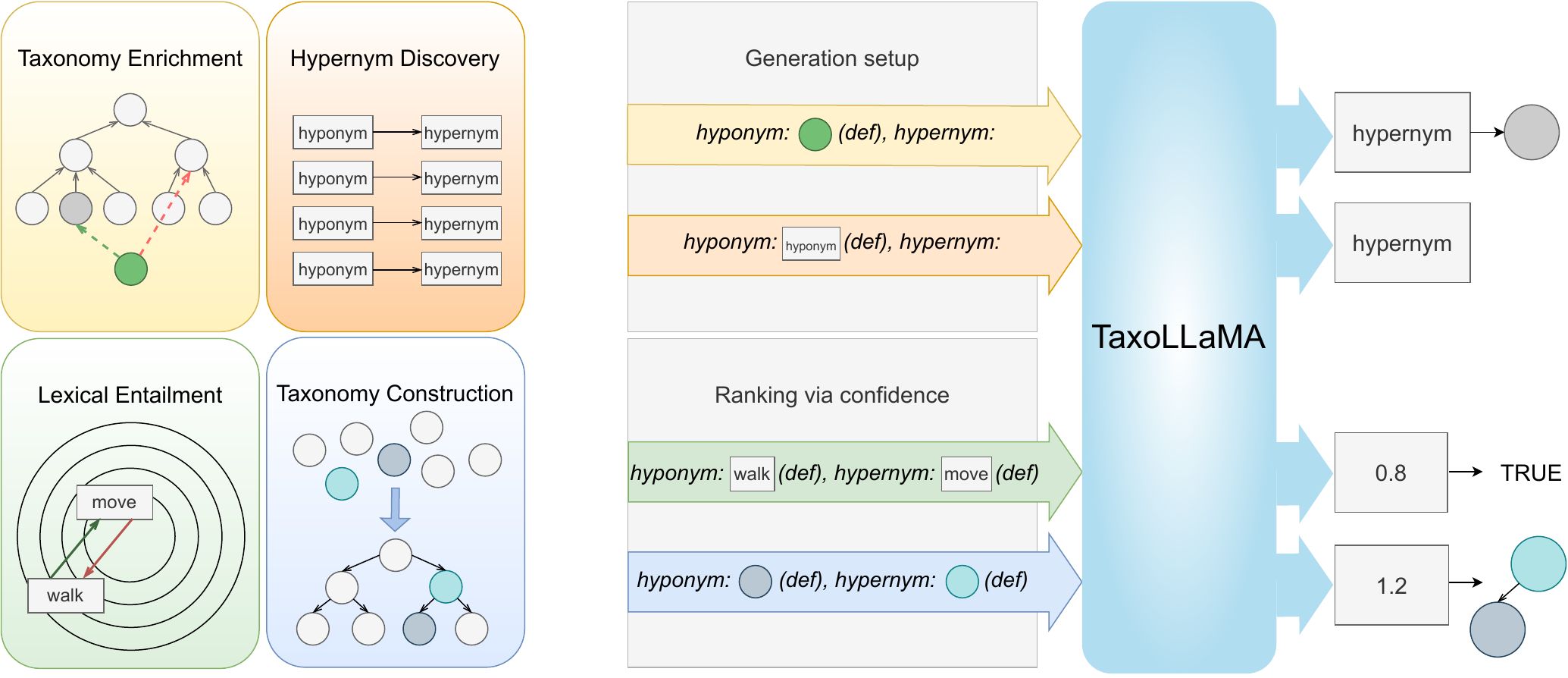}
\begin{subfigure}[b]{0.35\textwidth}
        \centering
        \caption{Lexical semantic tasks}
        \label{fig:pipeline_a}
    \end{subfigure}
\begin{subfigure}[b]{0.55\textwidth}
        \centering
        \caption{Generation and ranking pipelines for solution of various lexical semantic tasks }
        \label{fig:pipeline_b}
    \end{subfigure}
\vspace{-0.2cm}    \caption{Examples with input and output for each task are highlighted by color. Rectangle ``hypernym'' denotes a word generated by the model; circle means a node from the graph. Confidence score determines the existence of a relationship between the two nodes provided in the input.}
    \label{fig:pipeline}
    \vspace{-0.5cm}
\end{figure*}

The contributions of the paper are as follows:
\begin{itemize}
    \item We introduce the use of LLMs across various lexical semantic tasks via hypernym prediction and propose an appropriate taxonomy instruction tuning method that exploits \mbox{WordNet} for dataset sampling.
    \item We present \text{TaxoLLaMA}  -- a unified model designed to address a spectrum of lexical-sematic tasks achieving state-of-the-art (SOTA) results in 11 out of 16 tasks and securing the second rank in 4 tasks.
    \item We present an instructive dataset based on English WordNet-3.0 only for training a taxonomy-based LLM and collected definitions for input words in the Taxonomy Enrichment datasets and the Lexical Entailment datasets using Wikidata\footnote{\url{http://wikidata.org}} and ChatGPT\footnote{\url{https://chat.openai.com}}.
    \item We perform a detailed error analysis for all tasks using both manual and automatic approaches: e.g. we evaluate error patterns and model quality using ChatGPT.
\end{itemize}

\section{Related Work}

In this section, we briefly describe previous approaches to the lexical semantics tasks that we are experimenting with in the paper. %We also discuss approaches that both apply LLMs and WordNet.

\paragraph{Hypernym Discovery}

The Hypernym Discovery task involves predicting a list of hypernyms for a given hyponym (see example in Figure \ref{fig:pipeline_a}). The recent study introduces a taxonomy-adapted, fine-tuned T5 model \cite{nikishina-etal-2023-predicting}. Earlier models include the Recurrent Mapping Model (RMM) \cite{bai-etal-2021-hypernym}, which employs an Multilayer Perceptron (MLP) with residual connections and a contrastive-like loss. CRIM \cite{bernier-colborne-barriere-2018-crim}, distinguished as the best in SemEval, utilizes a similar MLP structure with a contrastive loss. The Hybrid model \cite{held-habash-2019-effectiveness} combines the k-Nearest Neighbor approach with Hearst patterns, while the 300-sparsans method \cite{berend-etal-2018-300} is an enhancement to the traditional word2vec approach.

\paragraph{Taxonomy Enrichment} 

This task is addressed in SemEval-2016 Task 14 \citep{jurgens2016semeval}, aiming to add a new word to the correct hypernym (node) in the given taxonomy. Numerous different architectures have been proposed to solve the task in recent years. TMN \citep{zhang2021taxonomy} exploits multiple scorers to find \textlangle hypernym, hyponym\textrangle{}  pairs for a given query concept. The TaxoEnrich \citep{jiang2022taxoenrich} employs two LSTMs \citep{staudemeyer2019understanding} to encode ancestors and descendants information. In addition, the TaxoExpan \citep{shen2020taxoexpan} uses Graph Neural Network (GNN)  \citep{scarselli2008graph} to predict whether the query is a child of an anchor concept. %while the GenTaxo model \citep{zeng2021enhancing} generates new concepts by fusing sentence- and graph-based representations.

\paragraph{Taxonomy Construction}

The taxonomy construction task aims to extract hypernym-hyponym relations between a given list of domain-specific terms and then construct a domain taxonomy based on them. The models for this task include Graph2Taxo \cite{shang-etal-2020-taxonomy}, which employs a sophisticated GNN architecture, LMScorer. RestrictMLM \cite{jain-espinosa-anke-2022-distilling} uses zero-shot RoBERTa or GPT2 for pair relationship scoring, differing in their use of MASK or next token probabilities. TAXI+ \cite{aly2019child} combines Hearst patterns with Poincaré embeddings for refinement of the existing approaches.

\paragraph{Lexical Entailment}

Lexical entailment is a classification task that identifies semantic relationships between phrase pairs. An example of the lexical entailment might be a hyponym ``cat'' which entails the existence of a hypernym ``animal''.

One of the recent lexical entailment models is LEAR \cite{Vulic:2018naacl} a fine-tuning method of transforming Euclidean space so that it reflects hyponymy-hypernymy relations. In SeVeN \cite{espinosa-anke-schockaert-2018-seven} relations between words are encoded. Pair2Vec \cite{joshi-etal-2019-pair2vec} and variant of GloVe introduced in \cite{jameel-etal-2018-unsupervised} use words' co-occurrence vectors and Pointwise Mutual Information. %RELATIVE \cite{ijcai2019p682} is a latent variable model that learns distributional relation vectors. 
GBL (``Global'' Entailment Graph) \citep{hosseini-etal-2018-learning} is GNN that utilizes ``local'' learning and CTX (``Contextual'' Entailment Graph) \cite{hosseini-etal-2021-open-domain} is the improvement of GBL with contextual link-prediction. \citet{mckenna-etal-2023-smoothing} proposes an etailment smoothing technique to the CTX model resulting in SOTA for the task. 

% взято из статьи на которой эти результаты есть
%GBL. The EG of Hosseini et al. (2018), introduces a “globalizing” graph-based method to improve the edges after “local” EG learning. CTX. The state-of-the-art contextualized EG of Hosseini et al. (2021), improves over GBL edges by augmenting local learning with a contextual link-prediction objective, before globalizing

% \paragraph{LLMs for other Lexical Semantic tasks}

% \paragraph{LLMs and WordNet}

% Inspired by \cite{nikishina-etal-2023-predicting,moskvoretskii-etal-2024-large-language}, we enhance the methodology by developing a large WordNet-based dataset and fine-tuning a Large Language Model with it. Enhanced with the WordNet knowledge such a model can compete with all of the models described above.

\section{Methodology}

This section outlines the process of data collection and the subsequent training of the model.

\subsection{Data Collection}

To create the dataset, we apply the algorithm presented by \citet{moskvoretskii-etal-2024-large-language}, focusing on hyponym-hypernym relationships only. We sample both nouns and verbs from the WordNet-3.0 graph. To prepare our training and validation sets, we randomly pick edges to form pairs of hyponym-hypernym, the motivation for precise choice is given in Section~\ref{sec:appendix_hypernym}. %We choose edges—not child nodes—because 
If a child node links to more than one hypernym, we count each link as a separate pair.
Additionally, we incorporate definitions for child nodes from WordNet to disambiguate the sense of the input word. As definitions may not be provided for some subtasks during inference (Lexical Entailment, MAG PSY, and MAG CS from Taxonomy Enrichment), we additionally generate definitions with ChatGPT for test sets that lack pre-defined explanations or take them from Wikidata. We use the web interface of ChatGPT 3.5 from February 2024 and the ``gpt-3.5-turbo'' model from the same period to generate definitions. The prompts for such requests and the statistics of the generated definitions are presented in the Appendix \ref{sec:chatgpt} in Examples \ref{chatgpt_dataset}-\ref{chatgpt_dataset2} and in Table \ref{tab:def_chatgpt}. This step is highly required:
the lack of definitions can reduce the performance of the model, as shown in \citet{moskvoretskii-etal-2024-large-language}. %This cannot, however, be attributed to hypernyms being mentioned in the definition: it was shown that extracting hypernyms from the definition performs worse than even simply using the word embeddings \cite{arefyev-etal-2020-word2vec}.

Below we show a training sample from our dataset used for instruction tuning of TaxoLLaMA. It comprises a system prompt describing the desired output (\ref{desired}) combined with an input word selected from WordNet, along with its definition (\ref{inputformat}), and the target (\ref{target}), which is the true hypernym of this input word, also sourced from WordNet:

\lb{desired}{
\textsf{\small[INST] <<SYS>> You are a helpful assistant. List all the possible words divided with a comma. Your answer should not include anything except the words divided by a comma <</SYS>>}}
%Then having collected pairs and definitions collected the prompt and target are the following:
\vspace{-0.6cm}
\lb{inputformat}{
\textsf{\small 
\textbf{hyponym}: tiger (large feline of forests in most of Asia having a tawny coat with black stripes) | \textbf{hypernyms}: [/INST]}} 
%}}
\vspace{-0.6cm}
\lb{target}{\textsf{\small 
big cat,}} 

The statistics of the generated datasets are provided in the next Subsection \ref{subsec:details} along with the setups they were created for.

\subsection{Training Details}\label{subsec:details}

We introduce two versions of our model: \text{TaxoLLaMA}, the model trained on the full WordNet-3.0 dataset for further community usage in lexical semantic tasks,  
%a broad and detailed model for different taxonomy tasks for 
and \text{TaxoLLaMA-bench}, designed for the benchmark tests. For this model, we make sure that the training set does not include any nodes from the test sets of those four tasks. The size of the training set for the first model is $44,772$ items, whereas the other model was finetuned with $36,775$ samples. The TaxoLLaMA-Verb model that we experiment with in Section \ref{sec:entailment} is fine-tuned exclusively on the verb sub-tree from WordNet, resulting in $7,712$ samples.
The finetuning procedure of our models is depicted in Figure \ref{fig:taxllama}. 

To train in this setup, we use the LLaMA-2 model with $7$ billion parameters \cite{llama2}. For better computational efficiency during training and inference, we quantize the model to 4 bits and fine-tune it using QLoRA \cite{dettmers2023qlora}, a full-precision LoRA adapter. During pre-training, we used a batch size of $32$ and a learning rate of $3e^{-4}$, applying a cosine annealing scheduler. Any further fine-tuning for different domains or languages was done with a batch size of 2 and a learning rate of $3e^{-4}$, without using schedulers. Other details are described in Appendix~\ref{sec:appendix_hyperparams}.

\begin{table*}[ht]
\centering

\begin{minipage}{0.9\textwidth}{
    \centering
    \resizebox{\textwidth}{!}{% Adjust the resizebox parameters as needed
    \begin{tabular}{l| ccccc}
    \toprule
          & 1A: English    & 2A: Medical    & 2B: Music & 1B: Italian & 1C: Spanish  \\ 
    \midrule
    %\midrule
    CRIM* \citep{bernier-colborne-barriere-2018-crim} &  36.10     &  54.64   & 60.93  & - & -   \\
    Hybrid* \citep{held-habash-2019-effectiveness}&  34.07 &  \underline{64.47} & \underline{77.24} & - & - \\
    RMM* \citep{bai-etal-2021-hypernym}& 39.07 & 54.89 & 74.75 & - & - \\
    T5 \citep{nikishina-etal-2023-predicting} & \underline{45.22} & 44.73 & 53.35 & 24.04 & 27.50 \\
    300-sparsans* \citep{berend-etal-2018-300}   & - & - & -   & \underline{25.14} & \underline{37.56} \\ 
    \midrule
    TaxoLLaMA zero-shot     & 38.05 & 43.09 & 42.7 & 1.95  & 2.21  \\
    TaxoLLaMA-bench zero-shot & 37.66 & 42.2  & 44.36 & 1.47  & 2.08 \\
    \midrule
     TaxoLLaMA     fine-tuned    & \textbf{54.39} & \textbf{77.32} & \textbf{80.6} & \textbf{51.58} & \textbf{57.44} \\
    TaxoLLaMA-bench fine-tuned   & \textbf{51.59} & \textbf{73.82} & \textbf{78.63} & \textbf{50.95} & \textbf{58.61} \\
    \bottomrule
    \end{tabular}
    }
    \caption{MRR performance on Hypernym Discovery. * refers to the systems that rely on the provided dataset only, without LLM pretraining or additional data being used. \textit{Zero-shot} is trained on the WordNet data only, without fine-tuning on the target dataset.}
    \label{table:hd_english}}
\end{minipage}\hfill
\end{table*}

\begin{figure*}[htbp]
\vspace{-0.5cm}
    \centering
    \begin{subfigure}[b]{0.48\textwidth}
        \centering
        \includegraphics[width=\textwidth]{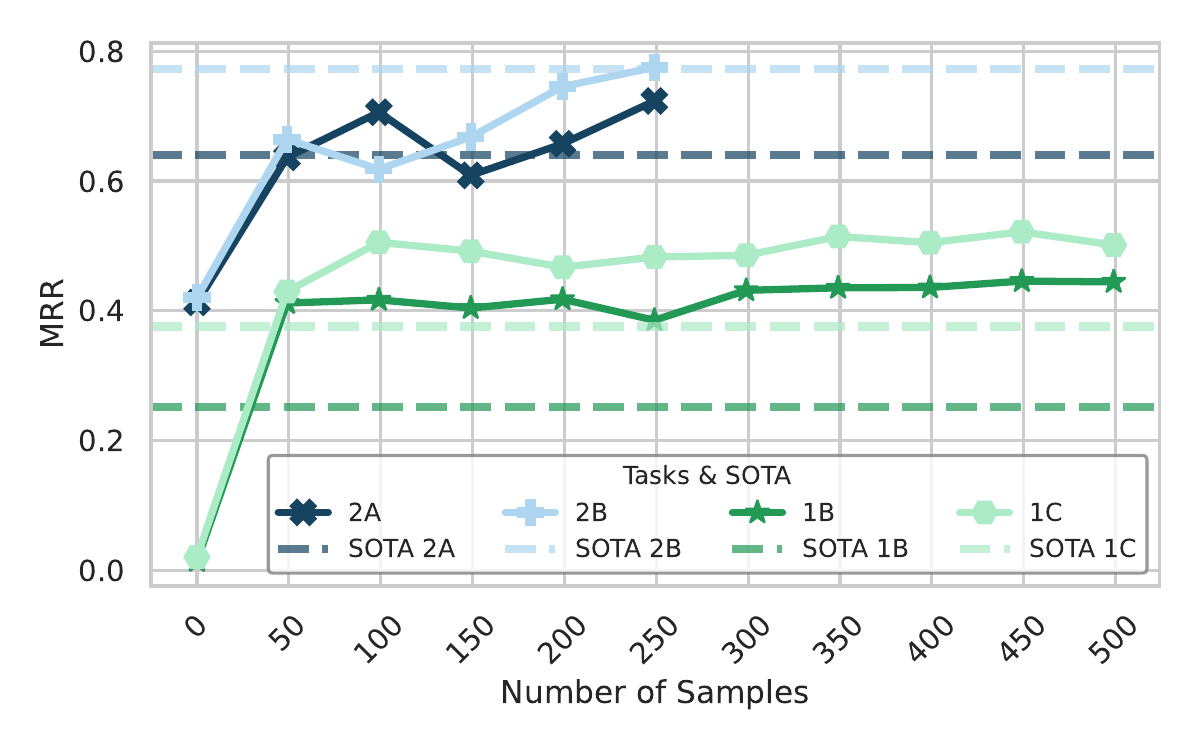}
        \caption{Fine-tuning}
        \label{fig:ablation_ft}
    \end{subfigure}
   % \hfill % Optional: might be needed to properly align the figures side by side
    \begin{subfigure}[b]{0.48\textwidth}
        \centering
        \includegraphics[width=\textwidth]{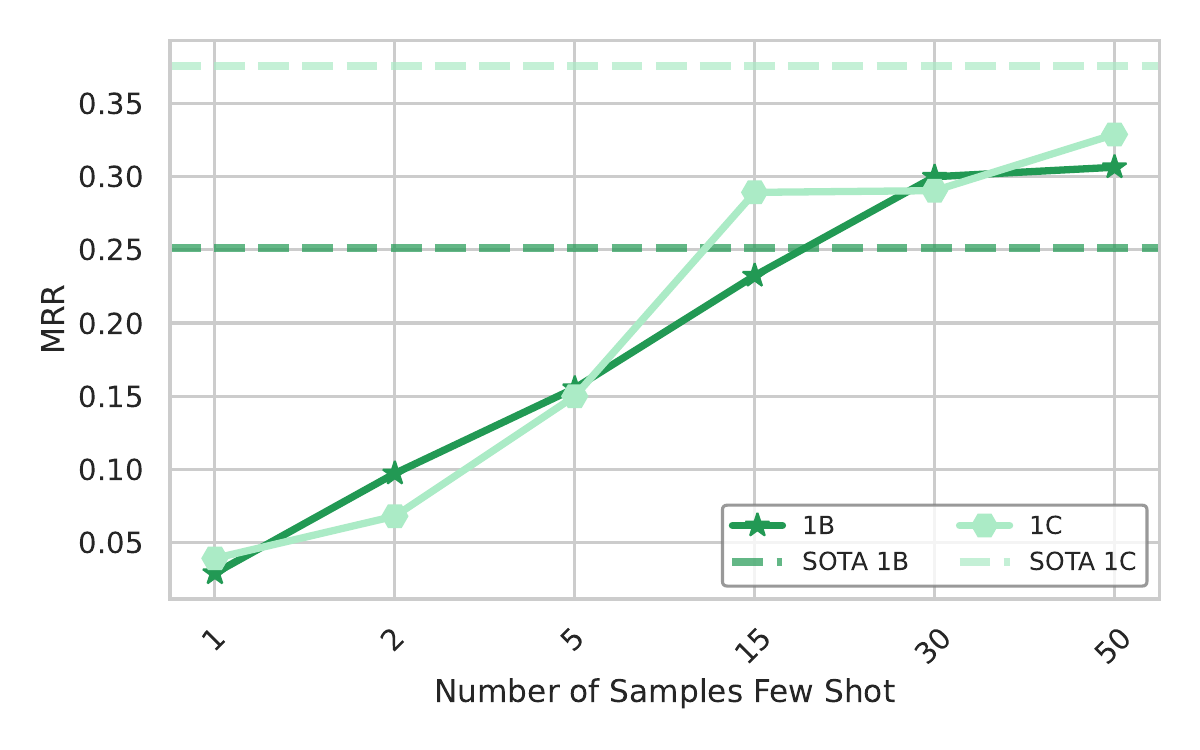}
        \caption{Few-shot learning}
        \label{fig:ablation_few_shot}
    \end{subfigure}
    \vspace{-0.2cm}\caption{Experiments for domain and language adaptation on the Hypernym Discovery datasets.}
    \label{fig:your-main-figure}
    \vspace{-0.5cm}
\end{figure*}

\subsection{Task Adaptation}
\label{sec:adaptation}

We propose two methods for adapting LLMs, fine-tuned with the WordNet instructive dataset. Different tasks interpret a taxonomy node in a different way and their understanding is reflected in Figure \ref{fig:pipeline}. For instance, Taxonomy Enrichment operates with synsets or synset names while Hypernym Discovery operates with lemmas (Figure \ref{fig:pipeline_a}). 
In Figure \ref{fig:pipeline_b}, there is no difference between ``TRUE'' and ``two-node connected''. However, they are depicted in such a way as to represent the expected outputs for two distinct tasks: Taxonomy Construction and Lexical Entailment. Taxonomy Construction focuses on creating a taxonomy graph from a list of nodes, essentially predicting the connections between them. On the other hand, lexical entailment involves determining whether a connection exists between two nodes.
We benefit from this task likeness because we can train one model that would be able to solve multiple lexical semantic tasks. Despite this, it's important to note that the tasks differ in how they interpret these relations.

We assume that Taxonomy-related tasks can be solved within two approaches from our pipeline.

\textbf{Generative approach} involves directly applying the same procedure as used in training. Given a hyponym, we use the model to generate a list of corresponding hypernyms. We apply this approach to the Hypernym Discovery and Taxonomy Enrichment datasets.

\textbf{Ranking approach} involves evaluating the hypernymy relation using perplexity: a lower score indicates a stronger relationship. Beyond assessing this relationship, we can also evaluate the hyponymy relation by simply reversing the hypernym and hyponym positions (this way we obtain reverse perplexity). The ratio between these two scores is a measure of confidence that we use for ranking. The lower the Confidence score, the higher the confidence of the model in the hypernymy relationship between the two constituents of a pair.

We apply this approach for the Taxonomy Construction and Lexical Entailment datasets with slight modifications that will be described in the respective sections \ref{sec:exp_texeval} and \ref{sec:exp_entail} in more detail.

\section{Experiments}

In this section, we assess the proposed methodology and the finetuned models, \text{TaxoLLaMA} and \text{TaxoLLaMA-bench}, on four lexical semantic tasks: Hypernym Discovery, Taxonomy Enrichment, Lexical Entailment, and Taxonomy Construction. We evaluate models in a zero-shot setting and after fine-tuning on the provided train sets for each task.

\subsection{Hypernym Discovery}

We test our model on the Hypernym Discovery task from SemEval-2018 \cite{camacho-collados-etal-2018-semeval} using our generative approach. This task features an English test set for general hypernyms and for two domain-specific ``Music'' and ``Medical'' sets, and general test sets for Italian and Spanish. Performance is measured using the Mean Reciprocal Rank (MRR) metric. We test a zero-shot approach, where the model is not tuned to the training datasets. The test set differs from WordNet and may involve multiple hops to hypernyms, and can also be applied to narrow domains.

\subsection{Taxonomy Enrichment}

Taxonomy Enrichment aims to identify the most appropriate placement for a missing node within a taxonomy. Continuing the approach of prior works \cite{zhang2021taxonomy, jiang2022taxoenrich}, the goal is framed as ranking nodes from the graph based on their likelihood of being the hypernym, where successfully placing the node means ranking its correct hypernyms at the top. In our setup, we use the generative approach described in Section \ref{sec:adaptation} and depicted in Figure \ref{fig:pipeline_b}.

The Taxonomy Enrichment benchmark encompasses the WordNet Noun, WordNet Verb, MAG-PSY, and MAG-CS datasets \cite{jiang2022taxoenrich,shen2020taxoexpan}. To ensure consistency, 1000 nodes from each dataset were sampled to match the test set from TaxoExpan \cite{shen2020taxoexpan}. 
%To evaluate our model performance, we leverage the generative approach, treating it similarly to hypernym discovery (see Figure \ref{fig:pipeline_a} for more detail). For the MAG-CS and MAG-PSY datasets, instead of ranking candidates from the provided taxonomy graph, we apply our dataset sampling strategy and fine-tune the model accordingly. 
Following \citet{jiang2022taxoenrich}, we consider scaled MRR \cite{Ying_2018} as the main metric, which is the regular MRR multiplied by 10 and averaged over all of a node's correct hypernyms.

\begin{table*}[ht]
\centering
\begin{minipage}{0.49\textwidth}
    \centering
    \resizebox{0.98\textwidth}{!}{
    \begin{tabular}{l|cccc}
    \toprule
     & MAG-CS & MAG-PSY & Noun & Verb \\ 
    \midrule
    TaxoExpan \citep{shen2020taxoexpan} & 19.3 & 44.1 & 39.0 & 32.5 \\
    GenTaxo \citep{zeng2021enhancing} & 23.9 & \underline{46.4} & 28.6 & 42.8 \\
    TMN \citep{zhang2021taxonomy} & 24.3 & 53.1 & 36.7 & 35.4 \\
    TaxoEnrich \citep{jiang2022taxoenrich} & \textbf{57.8} & \textbf{58.3} & \underline{44.2} & \underline{45.2} \\
    \midrule
    \text{TaxoLLaMA} zero-shot & 7.4 & 7.3 & n/a & n/a \\
    \text{TaxoLLaMA-bench} zero-shot & 8.5 & 6.6 & n/a & n/a \\
    \midrule
    \text{TaxoLLaMA} fine-tuned & 24.9 & 29.8 & \textbf{48.0} & \textbf{52.4} \\
    \text{TaxoLLaMA-bench} fine-tuned & \underline{30.2} & 31.4 & \textbf{45.9} & \textbf{51.9} \\
    %\text{TaxoLLaMA} ranker & 34.8 & - & - & - \\
    \bottomrule
    \end{tabular}
    }
    \caption{Scaled MRR Across Tasks for Taxonomy Enrichment. Here, ``n/a'' stands for ``not applicable'', as \text{TaxoLLaMA} has already seen WordNet data and its performance cannot be considered as zero-shot. \textit{Zero-shot} is trained on the WordNet data only, without fine-tuning on the target dataset.}
    \label{table:enrichment}
\end{minipage}\hfill
\begin{minipage}{0.49\textwidth}
\vspace{-0.5cm}
    \centering
    \resizebox{0.98\textwidth}{!}
    {
    \begin{tabular}{l|ccc}
    \toprule
     & S & E & F \\ 
    \midrule
    TexEval-2 best \citep{bordea-etal-2016-semeval}  & 31.3 & 30.0 &  \underline{36.01} \\
    TAXI+ \citep{aly2019child}& 41.4 & 30.9 & 34.1 \\
    Graph2Taxo pure \citep{shang-etal-2020-taxonomy}& 39.0 & 37.0 &  - \\
    Graph2Taxo best \citep{shang-etal-2020-taxonomy}& \textbf{47.0} & \underline{40.0} &  - \\
    LMScorer \citep{jain-espinosa-anke-2022-distilling} & 31.8 & 26.4 & 24.9 \\
    
    RestrictMLM \citep{jain-espinosa-anke-2022-distilling}  & 37.9 & 23.0 & 24.9 \\
    \midrule
    \text{TaxoLLaMA} & \underline{44.55} & \textbf{45.13} & \textbf{51.71} \\
    \text{TaxoLLaMA-bench} & \underline{42.36} & \textbf{44.82} & \textbf{51.18} \\
    \bottomrule
    \end{tabular}
    }
    \caption{F1 score for the  Taxonomy Construction Task. ``S'' stand for the (S)cience dataset, ``E'' for the (E)nvironment dataset, and ``F'' stands for the (F)ood domain dataset. \textit{Zero-shot} is trained on the WordNet data only, without fine-tuning on the target dataset.}

    \label{table:texeval}
\end{minipage}
\vspace{-0.4cm}
\end{table*}

\subsection{Taxonomy Construction}
\label{sec:exp_texeval}

This task aims to assemble a taxonomy given a list of nodes and a root. We employ datasets from TexEval-2 \cite{bordea-etal-2016-semeval} with ``Eurovoc science'', ``Eurovoc environment'' and ``WordNet food'' subtasks and the F1 measure for evaluation. % likewise in the original competition \cite{bordea-etal-2016-semeval}. 

We evaluate our model with the ranking approach applied to all node pairs.
Using this principle, we iteratively established a threshold below which pairs are considered to have a relationship. The threshold for the ``Food'' domain was set to $1.8$, for ``Environment'' to $4.6$, and for ``Science'' to $1.89$. To further refine the graph, we eliminate cycles by deleting the edge inside a cycle with
the highest perplexity. Additionally, we limit each node to a maximum of three hypernyms. For nodes associated with more than three hypernyms, only three with the lowest perplexity scores are retained.

\subsection{Lexical Entailment}
\label{sec:exp_entail}

Lexical Entailment aims at identifying semantic relationships between phrase pairs. Given a pair of words, the relation of entailment holds if there are some contexts in which one word can be substituted by the other, such that the meaning of the original word can be inferred from the new one.

We utilized the ANT entailment subset \cite{ANT} (a detailed enhancement of the Levy/Holt dataset \cite{holt2019probabilistic}) and Hyperlex benchmark \cite{vulić2017hyperlex} for our experiments.

\paragraph{ANT Dataset} This dataset contains pairs of sentences differing in one argument in syntactic structure (for example: ``The audience \textit{apploaded} the comedian'' and ``The audience \textit{observed} the comedian'', from Table 2 in \cite{ANT}). For these pairs, one of the relations is determined: antonymy, synonymy, directional entailment, or non-directional (which is reversed directional entailment) entailment. We treat the differing elements of the sentences as hypernym-hyponym pairs if the sentences are in one of the entailment relationships. To evaluate the entailment relations, we utilize the ratio of hypernym and hyponym ranking score, normalized via the L2 norm to represent the probability of entailment. 
For instance, we calculate the perplexity for ``move'' as a hypernym of ``walk'' (\(PPL_{m\rightarrow w}\)) and vice versa (\(PPL_{w\rightarrow m}\)). The ratio $\frac{PPL_{m\rightarrow w}}{PPL_{w\rightarrow m}}$ of these scores will thus indicate the model's confidence. 

\paragraph{HyperLex Dataset} This dataset focuses on the entailment for verbs and nouns, evaluating on a scale from 0 to 10. A score of 0 indicates no entailment, while 10 means strong entailment. The goal is to achieve the highest correlation with the gold-standard scores. For Hyperlex, we consider the ranking approach with no additional processing.

Previous methods generate embeddings and train a simple SVM on the Hyperlex training set. Fine-tuned models like RoBERTa demand substantial computational efforts and are tailored to the Hyperlex dataset. Compared to those prior studies, our zero-shot model uses perplexities directly as the predictions without a need for training. Therefore, a direct comparison might overlook the unique methodologies and resource implications, suggesting that each approach should be evaluated within its specific context.

\section{Results}
\label{sec:results}

This section describes the main results of generative and ranking setup experiments for all tasks.

\begin{table*}
    \centering
    \begin{minipage}{0.38\textwidth}
    \centering
    \resizebox{\textwidth}{!}{
    \begin{tabular}{l|cc}
    \toprule
     & AUC\textsubscript{N} & AP \\ 
    \midrule
    GBL \citep{hosseini-etal-2018-learning} & 3.79 & 58.36 \\
    CTX \citep{hosseini-etal-2021-open-domain} & 15.44 & 65.66 \\
    \midrule 
    GBL-P\textsubscript{K=4} \cite{mckenna-etal-2023-smoothing} & 13.91 & 64.71 \\
    CTX-P\textsubscript{K=4} \cite{mckenna-etal-2023-smoothing} & \textbf{25.86} & \underline{67.47} \\
    \midrule
    \text{TaxoLLaMA} zero-shot & 0.89 & 51.61 \\
    \text{TaxoLLaMA-bench} zero-shot & 2.82 & 54.24 \\
    \text{TaxoLLaMA-verb} zero-shot & \underline{19.28} & \textbf{69.51} \\
    \bottomrule
    \end{tabular}
    }
    \caption{Performance on the Lexical Entailment ANT dataset. \textit{Zero-shot} is trained on the WordNet data only, without fine-tuning on the target dataset.}
    \label{table:ANT}
    \end{minipage}
    \hspace{0.2cm}
    \centering
    \begin{minipage}{0.58\textwidth}
    \centering
    \resizebox{\textwidth}{!}{
    \begin{tabular}{llcc}
        \toprule
        Setting & Model & Lexical & Random \\
        \midrule
        \multirow{3}{*}{fine-tuned} & RoBERTa best \citep{Pitarch2023NoCG} & 79.4 & 82.8 \\
         & RoBERTa mean \citep{Pitarch2023NoCG}& 65.8 & 63.8 \\
         & LEAR \citep{Vulic:2018naacl}& 54.4 & 69.2 \\
        \Xhline{1pt} 
        \multirow{7}{*}{zero-shot} & Relative \citep{ijcai2019p682} & 54.3 & 58.4 \\
         & Pair2Vec \citep{joshi-etal-2019-pair2vec} & 33.4 & 54.3 \\
         & GRV SI \citep{jameel-etal-2018-unsupervised}& 48.3 & 55.4 \\
         & SeVeN \citep{espinosa-anke-schockaert-2018-seven}  & 46.9 & \textbf{62.7} \\
         & FastText & 43.9 & 54.3 \\
         \cmidrule(lr){2-4}
         & \text{TaxoLLaMA} & \textbf{70.2} & \underline{59.3} \\
        \bottomrule
        \end{tabular}
    }
    \caption{Spearman Correlation for lexical and random test subsets of Hyperlex benchmark.  \textit{Zero-shot} is trained on the WordNet data only, without fine-tuning on the target dataset.}
    \label{table:hyperlex}
    \end{minipage}
    \vspace{-0.5cm}
\end{table*}

\subsection{Hypernym Discovery}

The results for the English language in Table~\ref{table:hd_english}, indicate that both the fine-tuned \text{TaxoLLaMA} and \text{TaxoLLaMA-bench} outperform previous SOTA results by a large margin. While the zero-shot performance of our models may be lower than when fine-tuned, they still deliver comparable outcomes to previous results for general English tasks and do not fall far behind in domain-specific tasks, considering that previous approaches are all fine-tuned.

\paragraph{Multilingual Performance} For Italian and Spanish, the fine-tuned model surpasses previous SOTA results. 
We might assume the model's effectiveness in a multilingual setting, knowing that LLaMA-2 is initially multilingual and that previous finetuning was performed exclusively on English pairs. However, we observe that the zero-shot performance struggles to generate accurate hypernyms for languages other than English. It is worth mentioning that both Italian and Spanish data were not included in the instruction tuning dataset.

\paragraph{Zero-shot Performance} To investigate this zero-shot underperformance, we analyzed the effects of fine-tuning on both domains and languages, as shown in Figure~\ref{fig:ablation_ft}. It's clear that, except for task 2B, the model exceeds previous SOTA results with just 50 samples for fine-tuning. Additionally, the fluctuating scores highlight the model's sensitivity to the quality and nature of the training data.

\paragraph{Few-shot Performance} We also explored the few-shot learning approach for the Italian and Spanish languages to assess the model adaptability in an in-context learning environment, as shown in Figure~\ref{fig:ablation_few_shot}. 
The model surpassed previous SOTA benchmarks with a near-logarithmic pattern of improvement for the Italian language with 30 and 50 shots, yet not performing as well for Spanish. We attribute the suboptimal few-shot to the 4-bit quantization and its relatively small size. Smaller models typically exhibit lower performance across various benchmarks compared to their larger counterparts, as illustrated by the example of LLaMA-2 \cite{llama2}. Additionally, the capacity of smaller models or quantized models is also inferior compared to larger models, a finding corroborated by previous research \cite{wang-etal-2022-smash,frantar2023gptq,lin2023awq,egiazarian2024extreme}. The advantage gained from few-shot learning scenarios is less pronounced in quantized models compared to full-precision models. This observation has been specifically documented in the paper of \citet{lin2023awq}.
  %, as they might restrict in-context learning capabilities. 

%\input{tables/hypernym_prediction}

\subsection{Taxonomy Enrichment}

The results presented in Table~\ref{table:enrichment} show that our model surpasses all previous methods on the WordNet Noun and WordNet Verb datasets but does not perform as well as the current SOTA method on the more specialized MAG-CS and MAG-PSY taxonomies even after fine-tuning. We also notice that \text{TaxoLLaMA-bench}, having less data, unexpectedly performed better on the MAG datasets. To delve deeper into the reasons behind overall underperformance, we conducted a comprehensive error analysis, detailed in Section~\ref{err:enrichment}.

\subsection{Taxonomy Construction}

The results in Table~\ref{table:texeval} demonstrate that applying our method directly leads to SOTA performance on the ``Environment'' and ``Food'' datasets, and secures a second-place ranking for the ``Science'' dataset. Further analysis of the graphs generated through our modeling is provided in Section~\ref{err:texeval}.

\subsection{Lexical Entailment}\label{sec:entailment}

The results of \text{TaxoLLaMA} on the Lexical Entailment datasets surpassed our expectations.

\paragraph{Results on the ANT Dataset} From the results on the ANT dataset in Table~\ref{table:ANT}, we benchmark our models against prior SOTA performances. A notable finding is the obvious difference in performance between \text{TaxoLLaMA}, which is trained on both nouns and verbs, and \text{TaxoLLaMA-verb}, which focuses solely on verbs.

\text{TaxoLLaMA-verb} outperformes \text{TaxoLLaMA} in Lexical Entailment, suggesting difficulties in processing nouns and verbs simultaneously that might impede verb learning, possibly due to quantization and LORA adapter tuning constraints. This issue seems specific to the entailment task, as it does not emerge in other tasks, such as Taxonomy Enrichment, which also includes a verb dataset. This discrepancy could stem from metrics requiring precise normalized perplexity ranking.

Table \ref{table:ANT} shows that \text{TaxoLLaMA-verb} achieves SOTA performance on Average Precision and is second by normalized AUC. The comparison with previous SOTA results is skewed, as the best-performing models benefited from the use of additional Entailment Smoothing \cite{mckenna-etal-2023-smoothing} on top of the model. This technique has yet to be applied to our models, which might be a promising direction for future enhancements. %Excluding results that utilize Entailment Smoothing, our model would achieve SOTA status in both metrics, marking a substantial margin of improvement.

\paragraph{Results on the HyperLex Dataset} Table~\ref{table:hyperlex} demonstrates the superiority of our model over the previous SOTA in a zero-shot context for the ``Lexical'' subset and a second-place ranking for the ``Random'' subset. Contrary to the common trend where other models score higher on the random subset, our method does not follow this pattern, suggesting that the larger training size of the random subset benefits other methods more. Despite the straightforward zero-shot approach of our model, it still achieves notably high results. Future work could explore using this score as a meta-feature in task-specific models or adapting our entire model more closely to this task.

\begin{figure}[ht]
    \centering
    \vspace{-0.48cm}\includegraphics[width=0.499\textwidth]{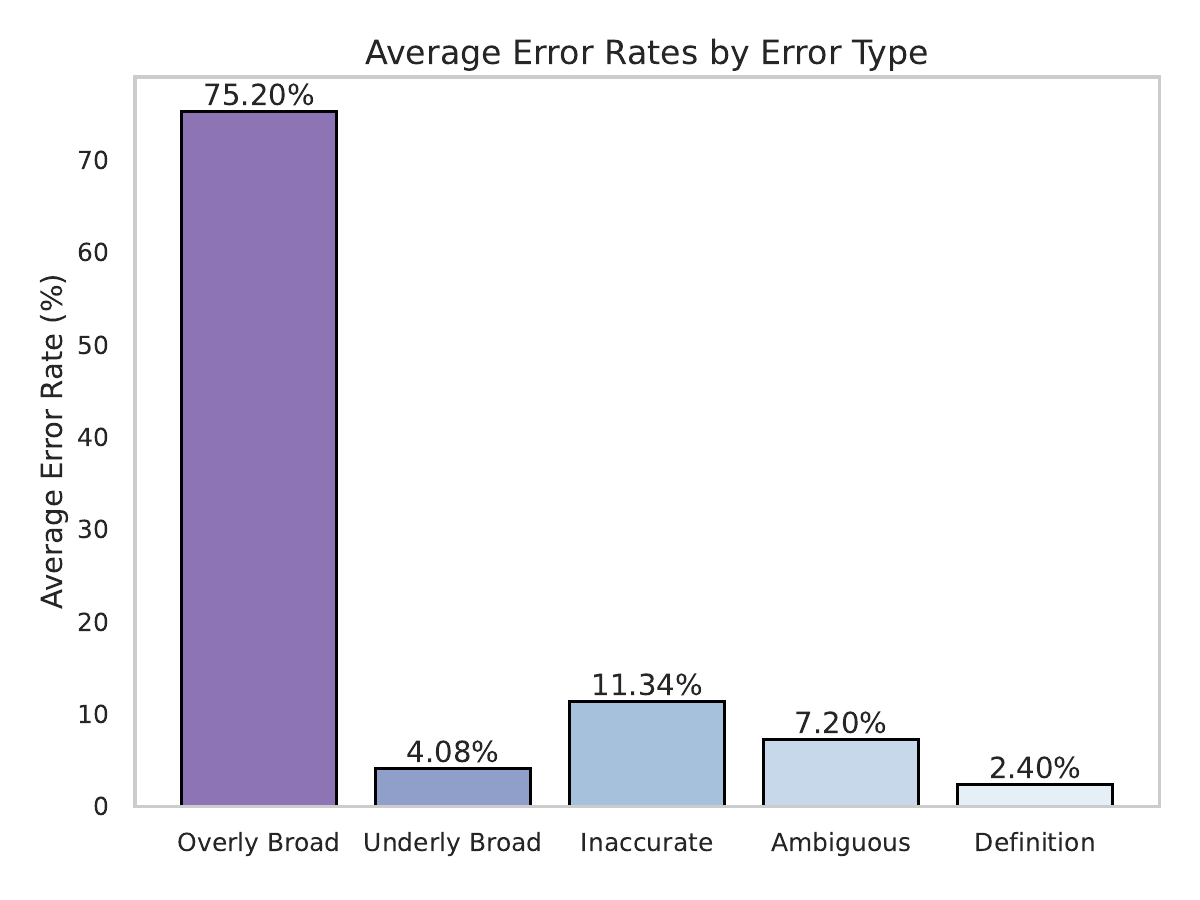}
    \vspace{-0.8cm}\caption{Average percentage of error types across Hypernym Discovery and Taxonomy Enrichment datasets.}
    \label{fig:error_type}
\vspace{-0.5cm}
\end{figure}

\section{Error Analysis}

In this section, we analyze the errors made by the \text{TaxoLLaMA} model, explore the reasons behind these inaccuracies, and suggest potential strategies for mitigation of LLMs applied to taxonomies.

\subsection{Hypernym Discovery and Taxonomy Enrichment} \label{err:enrichment}

%Here we inspect the errors made with the hypernym prediction setting. 
As we apply the same generative approach for both Hypernym Discovery and Taxonomy Enrichment we perform the joint error analysis. %We will focus more on the enrichment and especially MAG-CS and MAG-PSY as long as our model underperformed on those datasets.
We split the process into four steps: (i) manual analysis to identify the most common errors; (ii) automatic analysis of errors using ChatGPT; (iii) comparing and merging the most common errors identified; (iv) classification of the errors using ChatGPT. 

First, we take about 200 random examples from both Hypernym Discovery and Taxonomy Enrichment datasets and write explanations of why the model fails to generate the correct hypernym. The following four classes are identified: (i) predicted hypernyms are too broad; (ii) incorrect/irrelevant definition generated by ChatGPT; (iii) the model was unable to generate relevant candidates in the same semantic field; (iv) miscellaneous cases.

We also use the prompt in Example \ref{chatgpt_input} to ask ChatGPT to generate error types. The output is provided in Example \ref{chatgpt_output}; Table \ref{tab:chatgpt} summarizes all the error types generated during several runs. Then, we merge automatically and manually identified error types into the following classes:
%ChatGPT suggested about 12 different error types, supporting its viewpoint with examples from the input cases. The full list of error types, input prompt, and model output can be seen in Appendix \ref{sec:chatgpt} in Table \ref{tab:chatgpt} and Examples \ref{chatgpt_input} and \ref{chatgpt_output}. We noticed that ChatGPT tended to overgeneralize and treat very uncommon cases as a separate group. After comparing all manual and automatical classes, we merged them into the following error types:

\begin{enumerate}[noitemsep,topsep=0pt]
    \item \textbf{\text{Overly Broad Predictions}}: The model often generates predictions encompassing a broader concept than the true hypernym. 
    \item \textbf{\text{Underly Broad Predictions}}: Conversely, some predictions are too narrow and fail to capture the broader concept represented by the true hypernym.
    \item \textbf{\text{ Inaccurate Predictions}}: The model may predict words that are very semantically close to the true hypernym but struggles with fitting into the exact wording 
    \item \textbf{\text{Conceptual Ambiguity}}: The model may struggle with ambiguous input words or concepts, leading to incorrect predictions.
    \item \textbf{\text{Incorrect definitions}}: The model gets confused with the incorrect/inaccurate definition retrieved from external sources.
\end{enumerate}

\begin{figure}[t]
    \vspace{-0.48cm}\includegraphics[width=0.499\textwidth]{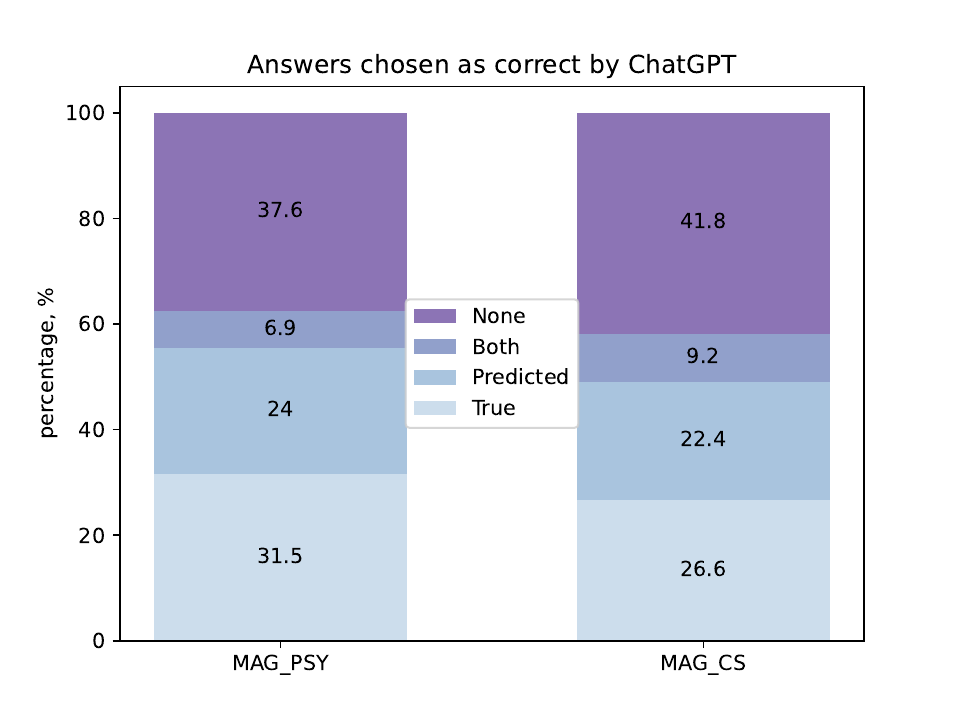}
    \vspace{-0.8cm}\caption{Automatic Evaluation of the MAG datasets with the ChatGPT model. \textit{``True''} denotes the number of gold answers that ChatGPT preferred over \text{TaxoLLaMA} answers; \textit{``Predicted''} is when ChatGPT preferred \text{TaxoLLaMA} output; \textit{``Both''} and \textit{``None''} options were also possible answers for ChatGPT.}
    \vspace{-0.5cm}
    \label{fig:gpt_eval}
    
\end{figure}

We used the prompt \ref{class_chatgpt} presented in Appendix \ref{sec:chatgpt} to classify incorrectly predicted instances. The results for each task and each dataset are presented in Appendix \ref{sec:erorr_type_analysis} in Table \ref{tab:errors_distribution} and in Figure \ref{fig:error_type} for average distribution. We also provide Table \ref{tab:examples} with an example for each error type. The most common issue (75\% of cases) is overly broad predicted concepts. It can be explained by the model adaptation to domain datasets that are richer than WordNet, like the ``Music'' and ``Medical'' domains.
%It can be partially explained by the slow adaptation of the model to new datasets that might have less specific concepts and structure than WordNet.
%In the ``Music'' and ``Medical'' domains, most errors are also classified as ``Overly Broad'' concepts within the fine-tuning sets. %, though such errors were relatively rare.
For Italian and Spanish, significant inaccuracies were attributed to grammatical complexities, due to the dataset limitations, linguistic intricacies, and lack of pre-training data.
%For Italian and Spanish, significant inaccuracies were attributed to grammatical complexities, with Spanish showing a unique level of ambiguity, because of the dataset limitations, linguistic intricacies, and lack of pre-training data.
Similarly, MAG datasets faced issues with specificity and ambiguity, which led to lower results of \text{TaxoLLaMA} compared with Wordnet-based ones, as shown in Table~\ref{table:enrichment}. %Therefore, we performed additional analysis of the output and true answers
%, largely due to unreliable definitions or confusion with common words.
%These challenges led to much lower results of \text{TaxoLLaMA}on the MAG datasets in comparison to the SOTA results on Wordnet-based ones, as shown in Table~\ref{table:enrichment}. Therefore, we performed additional analysis of the output and true answers.

% while the errors in the psychology domain were unexpectedly related to overly broad predictions, suggesting challenges in forming accurate, domain-specific relationships due to the field's less technical nature. 
%Those datasets exhibited considerable ambiguity, largely due to unreliable definitions or confusion with common words. 

% \begin{table*}[t]
% \centering
% \resizebox{0.98\textwidth}{!}{
% \begin{tabular}{lccccccccc}
% \toprule
%  & 1A: English & 2A: Medical & 2B: Music & 1B: Italian & 1C: Spanish & MAG-CS & MAG-PSY & Noun & Verb \\
% \midrule
% Error 1 & 361 & 75 & 88 & 169 & 134 & 468 & 610 & 417 & 342 \\
% Error 2 & 13 & 0 & 0 & 31 & 29 & 38 & 10 & 21 & 12 \\
% Error 3 & 47 & 1 & 0 & 86 & 33 & 93 & 80 & 31 & 98 \\
% Error 4 & 10 & 0 & 0 & 0 & 33 & 77 & 5 & 34 & 2 \\
% Error 5 & 67 & 4 & 0 & 23 & 44 & 28 & 9 & 9 & 12 \\
% \bottomrule
% \end{tabular}
% }
% \caption{Errors Distribution}
% \label{tab:errors_distribution}
% \end{table*}

%First, manual examination, revealed instances where the ground truth hypernyms appeared inaccurately placed, either missing a more precise intermediary node or, conversely, being overly specific. For example, ``olfactory toxicity in fish'' being classified as a hyponym of ``neuroscience'' seemed misaligned.
Manual examination of MAG taxonomies reveals misaligned instances, like ``olfactory toxicity in fish'' being classified as a hyponym of ``neuroscience''.
%to verify our hypothesis, we are the first to assess the accuracy of the predicted hypernyms using ChatGPT, a method increasingly employed in contemporary research \cite{rafailov2023direct}. 
Furthermore, we assess the accuracy of the predicted hypernyms using ChatGPT, inspired by contemporary research \cite{rafailov2023direct}. 
We provide the inputs, predicted nodes, and ground truth nodes to ChatGPT, asking for preference. As depicted in Figure~\ref{fig:gpt_eval}, ChatGPT mostly prefers neither of the answers and ground truth hypernyms only slightly more frequently than the predicted ones. The example of the input query is presented in Appendix \ref{sec:chatgpt} in Example \ref{class_chatgpt2}.

\begin{table}[t]
\centering
 \resizebox{0.5\textwidth}{!}{
\begin{tabular}{lccc}
\toprule
Metric & Science & Environment & Food \\ 
\midrule
\multicolumn{4}{c}{Original} \\
\midrule
\# Nodes & 125 & 261 & 1486 \\
\# Edges & 124 & 261 & 1533 \\
\midrule
\multicolumn{4}{c}{Constructed} \\
\midrule
\# Nodes & 78 & 216 & 1132 \\
\# Edges & 71 & 507 & 1372 \\
\# Nodes Missing & 48 & 45 & 354 \\
\# Weak Components & 8 & 5 & 51 \\
\# Nodes w/o original hypernym & 4 & 5 & 39 \\
\# Nodes w/o path to original hypernym & 29 & 70 & 308 \\
\# Nodes w/ path to original hypernym & 44 & 140 & 784 \\
Mean Distance to original hypernym & 1.02 & 1.15 & 1.06 \\ 
\bottomrule
\end{tabular}
}
\caption{Statistics of original graph and the constructed graph with highest F1 score. The lower part of the table corresponds to constructed graph statistics}
\label{table:graph-analysis}
\vspace{-0.5cm}
\end{table}

%We hope that this evaluation in the future will stimulate to the more accurate inspection of datasets for the taxonomy enrichment task and also considering LLMs as an evaluation tool.

We also evaluate the overlap between MAG datasets with WordNet data, discovering minimal correspondence. Only 5\% of the nodes from the entire graph are present in the WordNet graph, with just 2\% of edges for CS and 4\% for PSY matching. For 92\% of these, there is no path in the WordNet graph. Among the remaining connections, we see that 28\% in CS and 10\% in PSY represent cases where nodes are hypernyms of themselves. This also partially explains the low results of \text{TaxoLLaMA}, as MAG datasets greatly differ from the data used for  \text{TaxoLLaMA} training. 

Finally, we visualize the embeddings, revealing a notable discrepancy between predictions and ground truth in the MAG subsets—which was not seen with WordNet. ADetailed observations of this analysis are documented in Appendix~\ref{sec:appendix_tsne}.

\subsection{Taxonomy Construction} \label{err:texeval}

Our analysis of the predicted graphs across various domain datasets, based on statistics from Table~\ref{table:graph-analysis}, reveals consistent patterns. Generally, the gold standard graphs feature more edges, except in the environment domain. %, and consistently exhibit missing nodes. 
%Interestingly, the scarcity of missing parents suggests that 
The model often omits entire clusters of nodes rather than individual ones: about 30\% of nodes in the graph constructed with \text{TaxoLLaMA} lack a path to their actual parents, indicating they reside in separate components. %. This omission leads to numerous weak components within the graph, 
%which, if connected, could potentially degrade the overall score due to the absence of these nodes.

%We further examine the taxonomy construction errors with calculating some of the statistics.

Nevertheless, existing paths are of a rather high quality, suggesting the model is performing either very accurately or completely off-target. 
%At the same time, the paths that do exist among the remaining nodes are almost perfect. This dichotomy suggests that the model's predictions are either very accurate or completely off-target. 
%The low number of missing parents and the absence of certain graph paths imply that some  %
The model assigns high perplexity to certain paths which are further incorrectly excluded. %due to errors, certain graph paths, excluding them entirely, indicating 
This tendency indicates a particular challenge with concepts that are neither too specific nor too general but fall in the middle of the taxonomy.
%, where the model performance diminishes. This pattern of error could potentially impact the model effectiveness in other tasks. %highlighting a need for improved learning of mid-level taxonomy concepts. 

The nature of perplexity as a relative metric contributes to this issue, as some edges may not be created due to surpassing the perplexity threshold. Adjusting the threshold introduces incorrect edges, urging us to consider alternative approaches like using LLMs as embedders. %to enhance graph prediction accuracy.

\subsection{Lexical Entailment}

%Analyzing entailment is challenging due to complex metrics like normalized AUC, Average Precision, and Spearman correlation, alongside fluctuating normalized perplexities. 

Our examination of the ANT dataset showed it has nearly 3000 test samples but only 589 unique verbs. This means that errors on one verb could be replicated throughout the dataset. However, examining the overlap with WordNet revealed only 7 verbs were found in the same form. %, while others were different grammemes. 
Lemmatization increases the count to 338, but about 42\% of unique verbs still are not found in WordNet. No paths for the verbs presented in WordNet are found, which might have influenced model performance on the task.
Hyperlex demonstrates better statistics, with nearly half of the words being unique and 88\% found in WordNet. Only 27\% of pairs are presented in the taxonomy, and 99\% lack a connecting path. %, highlighting significant taxonomy alignment issues.

Perplexity-related errors show high values for polysemous pairs (e.g., ``spade is a type of card'') and low values for synonyms or paraphrases, indicating a semantic closeness but no hypernymy relation. This points to the model's struggle with lexical diversity and ambiguity, emphasizing the need for disambiguation abilities in entailment tasks. Additional analysis is available in Appendix~\ref{sec:appendix_corr}.

\section{Conclusion}

In this paper, we
introduce \text{TaxoLLaMA}--- an LLM finetuned on WordNet-3.0, capable of solving various
lexical semantic tasks via hypernym prediction. It achieved SOTA results in 11 out of 16 tasks and
securing the second position in 4 tasks.

Manual and ChatGPT-based error analysis shows that the most errors (75\%) are overly broad predicted concepts, due to overfitting to the idiosyncratic WordNet structure and inability to adapt to the target datasets. 
Experiments  showed that,  definitions greatly contribute to the final scores for Taxonomy Enrichment, similarly to \cite{moskvoretskii-etal-2024-large-language}, as they help to better disambiguate input words. Regarding error analysis, the most difficult datasets were MAGs~\cite{jiang2022taxoenrich}, as they greatly differ from the data used for training of our model.

% In future work, we plan to extend our research to other tasks and datasets and to enhance the finetuning with an additional loss function based on semantic similarity. We also noticed a need for improved learning of mid-level taxonomy concepts. Additionally, we will experiment with ``unlearning'' or selective forgetting \cite{DBLP:conf/emnlp/ChenY23} with the unified \text{TaxoLLaMA} model.

\section*{Limitations}

We find that the main limitations of our work  are as following:

\begin{itemize}
\item Dozens of large pre-trained generative models exist and we report results only on LLaMA-2. An alternative base LLM used could further improve the results. However, our experiments showed that LLaMA-2 showed decent performance on hypernymy prediction compared to other models. Moreover, our goal was also to provide a lightweight model that could be of further research with limited resources. Finally, the research focused on the LLM application and not on an exhaustive search of all LLM models. 

\item We did not apply the ``Ranking'' approach to the Taxonomy Enrichment dataset, which would be also possible, as finding the most appropriate node for the input word could also be seen as ranking. However, the first experiments showed lower results. %Therefore, we leave these experiments for further research.  % AP: I suggest to shorten, no need to say this in the limitation section.

\item Possible ``hypernymy hallucination'' may also be considered as a limitation: apart from the generalization capabilities the model may overpredict types, or even invent new words or semantic categories.

\item Another specificity of our model is its possible excessive focus on a single word sense, which may result in the inability to generate a wider variety of options.

\item We tried to be exhaustive, yet we possibly did not cover some taxonomy-related tasks. % , which we leave out of the scope of the research.  
\end{itemize}

\section*{Ethical Statement}

In our research, we employ advanced neural models like LLaMA-2, which have been pre-trained on a diverse corpus, including user-generated content. Although the creators of these models have endeavored to remove harmful or biased data, it is important to recognize that some biases may still persist in the model outputs. 

This acknowledgment does not undermine the validity of our methods. We have designed our techniques to be flexible, allowing them to be applied to alternative pre-trained models that have undergone more rigorous debiasing processes. To the best of our knowledge, aside from the challenge of mitigating inherent biases, our work does not raise any additional ethical concerns.

\section*{Acknowledgements}

This work was supported by the DFG through the project ``ACQuA: Answering Comparative Questions with Arguments'' (grants BI 1544/7- 1 and HA 5851/2- 1) as part of the priority program ``RATIO: Robust Argumentation Machines'' (SPP 1999). 

The work of Viktor Moskvoretskii was supported by Analytical center under the RF Government (subsidy agreement 000000D730321P5Q0002, Grant No. 70-2021-00145 02.11.2021).

\bibliography{anthology,custom}

\begin{thebibliography}{45}
\expandafter\ifx\csname natexlab\endcsname\relax\def\natexlab#1{#1}\fi

\bibitem[{Aly et~al.(2019)Aly, Acharya, Ossa, K{\"o}hn, Biemann, and Panchenko}]{aly2019child}
Rami Aly, Shantanu Acharya, Alexander Ossa, Arne K{\"o}hn, Chris Biemann, and Alexander Panchenko. 2019.
\newblock \href {https://doi.org/10.18653/v1/P19-1474} {Every child should have parents: A taxonomy refinement algorithm based on hyperbolic term embeddings}.
\newblock In \emph{Proceedings of the 57th Annual Meeting of the Association for Computational Linguistics}, pages 4811--4817, Florence, Italy. Association for Computational Linguistics.

\bibitem[{Bai et~al.(2021)Bai, Zhang, Kong, Chen, and Mao}]{bai-etal-2021-hypernym}
Yuhang Bai, Richong Zhang, Fanshuang Kong, Junfan Chen, and Yongyi Mao. 2021.
\newblock \href {https://doi.org/10.18653/v1/2021.findings-acl.257} {Hypernym discovery via a recurrent mapping model}.
\newblock In \emph{Findings of the Association for Computational Linguistics: ACL-IJCNLP 2021}, pages 2912--2921, Online. Association for Computational Linguistics.

\bibitem[{Berend et~al.(2018)Berend, Makrai, and F{\"o}ldi{\'a}k}]{berend-etal-2018-300}
G{\'a}bor Berend, M{\'a}rton Makrai, and P{\'e}ter F{\"o}ldi{\'a}k. 2018.
\newblock \href {https://doi.org/10.18653/v1/S18-1152} {300-sparsans at {S}em{E}val-2018 task 9: Hypernymy as interaction of sparse attributes}.
\newblock In \emph{Proceedings of the 12th International Workshop on Semantic Evaluation}, pages 928--934, New Orleans, Louisiana. Association for Computational Linguistics.

\bibitem[{Bernier-Colborne and Barri{\`e}re(2018)}]{bernier-colborne-barriere-2018-crim}
Gabriel Bernier-Colborne and Caroline Barri{\`e}re. 2018.
\newblock \href {https://doi.org/10.18653/v1/S18-1116} {{CRIM} at {S}em{E}val-2018 task 9: A hybrid approach to hypernym discovery}.
\newblock In \emph{Proceedings of the 12th International Workshop on Semantic Evaluation}, pages 725--731, New Orleans, Louisiana. Association for Computational Linguistics.

\bibitem[{Bordea et~al.(2016)Bordea, Lefever, and Buitelaar}]{bordea-etal-2016-semeval}
Georgeta Bordea, Els Lefever, and Paul Buitelaar. 2016.
\newblock \href {https://doi.org/10.18653/v1/S16-1168} {{S}em{E}val-2016 task 13: Taxonomy extraction evaluation ({TE}x{E}val-2)}.
\newblock In \emph{Proceedings of the 10th International Workshop on Semantic Evaluation ({S}em{E}val-2016)}, pages 1081--1091, San Diego, California. Association for Computational Linguistics.

\bibitem[{Camacho-Collados et~al.(2018)Camacho-Collados, Delli~Bovi, Espinosa-Anke, Oramas, Pasini, Santus, Shwartz, Navigli, and Saggion}]{camacho-collados-etal-2018-semeval}
Jose Camacho-Collados, Claudio Delli~Bovi, Luis Espinosa-Anke, Sergio Oramas, Tommaso Pasini, Enrico Santus, Vered Shwartz, Roberto Navigli, and Horacio Saggion. 2018.
\newblock \href {https://doi.org/10.18653/v1/S18-1115} {{S}em{E}val-2018 task 9: Hypernym discovery}.
\newblock In \emph{Proceedings of the 12th International Workshop on Semantic Evaluation}, pages 712--724, New Orleans, Louisiana. Association for Computational Linguistics.

\bibitem[{Camacho-Collados et~al.(2019)Camacho-Collados, Espinosa-Anke, Jameel, and Schockaert}]{ijcai2019p682}
Jose Camacho-Collados, Luis Espinosa-Anke, Shoaib Jameel, and Steven Schockaert. 2019.
\newblock \href {https://doi.org/10.24963/ijcai.2019/682} {A latent variable model for learning distributional relation vectors}.
\newblock In \emph{Proceedings of the Twenty-Eighth International Joint Conference on Artificial Intelligence, {IJCAI-19}}, pages 4911--4917. International Joint Conferences on Artificial Intelligence Organization.

\bibitem[{Dettmers et~al.(2023)Dettmers, Pagnoni, Holtzman, and Zettlemoyer}]{dettmers2023qlora}
Tim Dettmers, Artidoro Pagnoni, Ari Holtzman, and Luke Zettlemoyer. 2023.
\newblock \href {https://openreview.net/forum?id=OUIFPHEgJU} {{QL}o{RA}: Efficient finetuning of quantized {LLM}s}.
\newblock In \emph{Thirty-seventh Conference on Neural Information Processing Systems}.

\bibitem[{Dong et~al.(2023)Dong, Li, Dai, Zheng, Wu, Chang, Sun, Xu, Li, and Sui}]{dong2023survey}
Qingxiu Dong, Lei Li, Damai Dai, Ce~Zheng, Zhiyong Wu, Baobao Chang, Xu~Sun, Jingjing Xu, Lei Li, and Zhifang Sui. 2023.
\newblock \href {http://arxiv.org/abs/2301.00234} {A survey on in-context learning}.

\bibitem[{Egiazarian et~al.(2024)Egiazarian, Panferov, Kuznedelev, Frantar, Babenko, and Alistarh}]{egiazarian2024extreme}
Vage Egiazarian, Andrei Panferov, Denis Kuznedelev, Elias Frantar, Artem Babenko, and Dan Alistarh. 2024.
\newblock \href {http://arxiv.org/abs/2401.06118} {Extreme compression of large language models via additive quantization}.

\bibitem[{Espinosa-Anke and Schockaert(2018)}]{espinosa-anke-schockaert-2018-seven}
Luis Espinosa-Anke and Steven Schockaert. 2018.
\newblock \href {https://aclanthology.org/C18-1225} {{S}e{V}e{N}: Augmenting word embeddings with unsupervised relation vectors}.
\newblock In \emph{Proceedings of the 27th International Conference on Computational Linguistics}, pages 2653--2665, Santa Fe, New Mexico, USA. Association for Computational Linguistics.

\bibitem[{Frantar et~al.(2023)Frantar, Ashkboos, Hoefler, and Alistarh}]{frantar2023gptq}
Elias Frantar, Saleh Ashkboos, Torsten Hoefler, and Dan Alistarh. 2023.
\newblock \href {http://arxiv.org/abs/2210.17323} {{GPTQ}: Accurate post-training quantization for generative pre-trained transformers}.

\bibitem[{Guillou and de~Vroe(2023)}]{ANT}
Liane Guillou and Sander~Bijl de~Vroe. 2023.
\newblock \href {https://github.com/lguillou/ant} {{ANT} dataset}.

\bibitem[{Held and Habash(2019)}]{held-habash-2019-effectiveness}
William Held and Nizar Habash. 2019.
\newblock \href {https://doi.org/10.18653/v1/P19-1327} {The effectiveness of simple hybrid systems for hypernym discovery}.
\newblock In \emph{Proceedings of the 57th Annual Meeting of the Association for Computational Linguistics}, pages 3362--3367, Florence, Italy. Association for Computational Linguistics.

\bibitem[{Holt(2019)}]{holt2019probabilistic}
Xavier Holt. 2019.
\newblock \href {http://arxiv.org/abs/1907.12048} {Probabilistic models of relational implication}.

\bibitem[{Hosseini et~al.(2018)Hosseini, Chambers, Reddy, Holt, Cohen, Johnson, and Steedman}]{hosseini-etal-2018-learning}
Mohammad~Javad Hosseini, Nathanael Chambers, Siva Reddy, Xavier~R. Holt, Shay~B. Cohen, Mark Johnson, and Mark Steedman. 2018.
\newblock \href {https://doi.org/10.1162/tacl_a_00250} {Learning typed entailment graphs with global soft constraints}.
\newblock \emph{Transactions of the Association for Computational Linguistics}, 6:703--717.

\bibitem[{Hosseini et~al.(2021)Hosseini, Cohen, Johnson, and Steedman}]{hosseini-etal-2021-open-domain}
Mohammad~Javad Hosseini, Shay~B. Cohen, Mark Johnson, and Mark Steedman. 2021.
\newblock \href {https://doi.org/10.18653/v1/2021.findings-emnlp.238} {Open-domain contextual link prediction and its complementarity with entailment graphs}.
\newblock In \emph{Findings of the Association for Computational Linguistics: EMNLP 2021}, pages 2790--2802, Punta Cana, Dominican Republic. Association for Computational Linguistics.

\bibitem[{Hu et~al.(2022)Hu, Shen, Wallis, Allen-Zhu, Li, Wang, Wang, and Chen}]{hu2021lora}
Edward~J Hu, Yelong Shen, Phillip Wallis, Zeyuan Allen-Zhu, Yuanzhi Li, Shean Wang, Lu~Wang, and Weizhu Chen. 2022.
\newblock \href {https://openreview.net/forum?id=nZeVKeeFYf9} {Lo{RA}: Low-rank adaptation of large language models}.
\newblock In \emph{International Conference on Learning Representations}.

\bibitem[{Jain and Espinosa~Anke(2022)}]{jain-espinosa-anke-2022-distilling}
Devansh Jain and Luis Espinosa~Anke. 2022.
\newblock \href {https://doi.org/10.18653/v1/2022.starsem-1.13} {Distilling hypernymy relations from language models: On the effectiveness of zero-shot taxonomy induction}.
\newblock In \emph{Proceedings of the 11th Joint Conference on Lexical and Computational Semantics}, pages 151--156, Seattle, Washington. Association for Computational Linguistics.

\bibitem[{Jameel et~al.(2018)Jameel, Bouraoui, and Schockaert}]{jameel-etal-2018-unsupervised}
Shoaib Jameel, Zied Bouraoui, and Steven Schockaert. 2018.
\newblock \href {https://doi.org/10.18653/v1/P18-1003} {Unsupervised learning of distributional relation vectors}.
\newblock In \emph{Proceedings of the 56th Annual Meeting of the Association for Computational Linguistics (Volume 1: Long Papers)}, pages 23--33, Melbourne, Australia. Association for Computational Linguistics.

\bibitem[{Jiang et~al.(2022)Jiang, Song, Zhang, and Han}]{jiang2022taxoenrich}
Minhao Jiang, Xiangchen Song, Jieyu Zhang, and Jiawei Han. 2022.
\newblock \href {https://doi.org/10.1145/3485447.3511935} {Taxo{E}nrich: Self-supervised taxonomy completion via structure-semantic representations}.
\newblock In \emph{Proceedings of the ACM Web Conference 2022}, WWW '22, page 925–934, New York, NY, USA. Association for Computing Machinery.

\bibitem[{Joshi et~al.(2019)Joshi, Choi, Levy, Weld, and Zettlemoyer}]{joshi-etal-2019-pair2vec}
Mandar Joshi, Eunsol Choi, Omer Levy, Daniel Weld, and Luke Zettlemoyer. 2019.
\newblock \href {https://www.aclweb.org/anthology/N19-1362} {Pair2vec: Compositional word-pair embeddings for cross-sentence inference}.
\newblock In \emph{Proceedings of the 2019 Conference of the North {A}merican Chapter of the Association for Computational Linguistics: Human Language Technologies, Volume 1 (Long and Short Papers)}, pages 3597--3608, Minneapolis, Minnesota. Association for Computational Linguistics.

\bibitem[{Jurgens and Pilehvar(2016)}]{jurgens2016semeval}
David Jurgens and Mohammad~Taher Pilehvar. 2016.
\newblock \href {https://doi.org/10.18653/v1/S16-1169} {{S}em{E}val-2016 task 14: Semantic taxonomy enrichment}.
\newblock In \emph{Proceedings of the 10th International Workshop on Semantic Evaluation ({S}em{E}val-2016)}, pages 1092--1102, San Diego, California. Association for Computational Linguistics.

\bibitem[{Kauf et~al.(2023)Kauf, Ivanova, Rambelli, Chersoni, She, Chowdhury, Fedorenko, and Lenci}]{kauf2023event}
Carina Kauf, Anna~A Ivanova, Giulia Rambelli, Emmanuele Chersoni, Jingyuan~Selena She, Zawad Chowdhury, Evelina Fedorenko, and Alessandro Lenci. 2023.
\newblock \href {https://doi.org/10.1111/cogs.13386} {Event knowledge in large language models: the gap between the impossible and the unlikely}.
\newblock \emph{Cognitive Science}, 47(11):e13386.

\bibitem[{Lin et~al.(2024)Lin, Tang, Tang, Yang, Chen, Wang, Xiao, Dang, Gan, and Han}]{lin2023awq}
Ji~Lin, Jiaming Tang, Haotian Tang, Shang Yang, Wei-Ming Chen, Wei-Chen Wang, Guangxuan Xiao, Xingyu Dang, Chuang Gan, and Song Han. 2024.
\newblock \href {https://doi.org/10.48550/arXiv.2106.09685} {{AWQ}: Activation-aware weight quantization for llm compression and acceleration}.
\newblock In \emph{MLSys}.

\bibitem[{McKenna et~al.(2023)McKenna, Li, Johnson, and Steedman}]{mckenna-etal-2023-smoothing}
Nick McKenna, Tianyi Li, Mark Johnson, and Mark Steedman. 2023.
\newblock \href {https://aclanthology.org/2023.ijcnlp-main.37} {Smoothing entailment graphs with language models}.
\newblock In \emph{Proceedings of the 13th International Joint Conference on Natural Language Processing and the 3rd Conference of the Asia-Pacific Chapter of the Association for Computational Linguistics (Volume 1: Long Papers)}, pages 551--563, Nusa Dua, Bali. Association for Computational Linguistics.

\bibitem[{Miller(1998)}]{miller1998wordnet}
George~A Miller. 1998.
\newblock \emph{WordNet: An electronic lexical database}.
\newblock MIT press.

\bibitem[{Moskvoretskii et~al.(2024)Moskvoretskii, Panchenko, and Nikishina}]{moskvoretskii-etal-2024-large-language}
Viktor Moskvoretskii, Alexander Panchenko, and Irina Nikishina. 2024.
\newblock \href {https://aclanthology.org/2024.lrec-main.133} {Are large language models good at lexical semantics? a case of taxonomy learning}.
\newblock In \emph{Proceedings of the 2024 Joint International Conference on Computational Linguistics, Language Resources and Evaluation (LREC-COLING 2024)}, pages 1498--1510, Torino, Italia. ELRA and ICCL.

\bibitem[{Nikishina et~al.(2023)Nikishina, Chernomorchenko, Demidova, Panchenko, and Biemann}]{nikishina-etal-2023-predicting}
Irina Nikishina, Polina Chernomorchenko, Anastasiia Demidova, Alexander Panchenko, and Chris Biemann. 2023.
\newblock \href {https://aclanthology.org/2023.findings-ijcnlp.12} {Predicting terms in {IS}-a relations with pre-trained transformers}.
\newblock In \emph{Findings of the Association for Computational Linguistics: IJCNLP-AACL 2023 (Findings)}, pages 134--148, Nusa Dua, Bali. Association for Computational Linguistics.

\bibitem[{Pitarch et~al.(2023)Pitarch, Bernad, Dranca, Bobed~Lisbona, and Gracia}]{Pitarch2023NoCG}
Lucia Pitarch, Jordi Bernad, Lacramioara Dranca, Carlos Bobed~Lisbona, and Jorge Gracia. 2023.
\newblock \href {https://doi.org/10.18653/v1/2023.acl-long.308} {No clues good clues: out of context lexical relation classification}.
\newblock In \emph{Proceedings of the 61st Annual Meeting of the Association for Computational Linguistics (Volume 1: Long Papers)}, pages 5607--5625, Toronto, Canada. Association for Computational Linguistics.

\bibitem[{Rafailov et~al.(2023)Rafailov, Sharma, Mitchell, Manning, Ermon, and Finn}]{rafailov2023direct}
Rafael Rafailov, Archit Sharma, Eric Mitchell, Christopher~D. Manning, Stefano Ermon, and Chelsea Finn. 2023.
\newblock \href {https://arxiv.org/abs/2305.18290} {Direct preference optimization: Your language model is secretly a reward model}.
\newblock In \emph{NeurIPS}.

\bibitem[{Reimers and Gurevych(2019)}]{reimers2019sentencebert}
Nils Reimers and Iryna Gurevych. 2019.
\newblock \href {https://doi.org/10.18653/v1/D19-1410} {Sentence-{BERT}: Sentence embeddings using {S}iamese {BERT}-networks}.
\newblock In \emph{Proceedings of the 2019 Conference on Empirical Methods in Natural Language Processing and the 9th International Joint Conference on Natural Language Processing (EMNLP-IJCNLP)}, pages 3982--3992, Hong Kong, China. Association for Computational Linguistics.

\bibitem[{Scarselli et~al.(2008)Scarselli, Gori, Tsoi, Hagenbuchner, and Monfardini}]{scarselli2008graph}
Franco Scarselli, Marco Gori, Ah~Chung Tsoi, Markus Hagenbuchner, and Gabriele Monfardini. 2008.
\newblock \href {https://doi.org/10.1109/TNN.2008.2005605} {The graph neural network model}.
\newblock \emph{IEEE transactions on neural networks}, 20(1):61--80.

\bibitem[{Shang et~al.(2020)Shang, Dash, Chowdhury, Mihindukulasooriya, and Gliozzo}]{shang-etal-2020-taxonomy}
Chao Shang, Sarthak Dash, Md. Faisal~Mahbub Chowdhury, Nandana Mihindukulasooriya, and Alfio Gliozzo. 2020.
\newblock \href {https://doi.org/10.18653/v1/2020.acl-main.199} {Taxonomy construction of unseen domains via graph-based cross-domain knowledge transfer}.
\newblock In \emph{Proceedings of the 58th Annual Meeting of the Association for Computational Linguistics}, pages 2198--2208, Online. Association for Computational Linguistics.

\bibitem[{Shen et~al.(2020)Shen, Shen, Xiong, Wang, Wang, and Han}]{shen2020taxoexpan}
Jiaming Shen, Zhihong Shen, Chenyan Xiong, Chi Wang, Kuansan Wang, and Jiawei Han. 2020.
\newblock \href {https://arxiv.org/abs/2001.09522} {Taxo{E}xpan: Self-supervised taxonomy expansion with position-enhanced graph neural network}.
\newblock In \emph{Proceedings of The Web Conference 2020}, pages 486--497.

\bibitem[{Staudemeyer and Morris(2019)}]{staudemeyer2019understanding}
Ralf~C Staudemeyer and Eric~Rothstein Morris. 2019.
\newblock \href {https://doi.org/10.48550/arXiv.1909.09586} {Understanding lstm--a tutorial into long short-term memory recurrent neural networks}.
\newblock \emph{arXiv preprint arXiv:1909.09586}.

\bibitem[{Sun et~al.(2023)Sun, Xu, Zha, Liu, and Dong}]{sun2023head}
Kai Sun, Yifan~Ethan Xu, Hanwen Zha, Yue Liu, and Xin~Luna Dong. 2023.
\newblock \href {https://arxiv.org/abs/2308.10168} {Head-to-tail: How knowledgeable are large language models (llm)? aka will llms replace knowledge graphs?}
\newblock \emph{arXiv preprint arXiv:2308.10168}.

\bibitem[{Tang et~al.(2023)Tang, Zhang, Lin, and Ture}]{tang2023llamas}
Raphael Tang, Xinyu Zhang, Jimmy Lin, and Ferhan Ture. 2023.
\newblock \href {https://doi.org/10.48550/arXiv.2311.18812} {What do llamas really think? revealing preference biases in language model representations}.
\newblock \emph{arXiv preprint arXiv:2311.18812}.

\bibitem[{Touvron et~al.(2023)Touvron, Martin, Stone, Albert, Almahairi, Babaei, Bashlykov, Batra, Bhargava, Bhosale, Bikel, Blecher, Canton{-}Ferrer, Chen, Cucurull, Esiobu, Fernandes, Fu, Fu, Fuller, Gao, Goswami, Goyal, Hartshorn, Hosseini, Hou, Inan, Kardas, Kerkez, Khabsa, Kloumann, Korenev, Koura, Lachaux, Lavril, Lee, Liskovich, Lu, Mao, Martinet, Mihaylov, Mishra, Molybog, Nie, Poulton, Reizenstein, Rungta, Saladi, Schelten, Silva, Smith, Subramanian, Tan, Tang, Taylor, Williams, Kuan, Xu, Yan, Zarov, Zhang, Fan, Kambadur, Narang, Rodriguez, Stojnic, Edunov, and Scialom}]{llama2}
Hugo Touvron, Louis Martin, Kevin Stone, Peter Albert, Amjad Almahairi, Yasmine Babaei, Nikolay Bashlykov, Soumya Batra, Prajjwal Bhargava, Shruti Bhosale, Dan Bikel, Lukas Blecher, Cristian Canton{-}Ferrer, Moya Chen, Guillem Cucurull, David Esiobu, Jude Fernandes, Jeremy Fu, Wenyin Fu, Brian Fuller, Cynthia Gao, Vedanuj Goswami, Naman Goyal, Anthony Hartshorn, Saghar Hosseini, Rui Hou, Hakan Inan, Marcin Kardas, Viktor Kerkez, Madian Khabsa, Isabel Kloumann, Artem Korenev, Punit~Singh Koura, Marie{-}Anne Lachaux, Thibaut Lavril, Jenya Lee, Diana Liskovich, Yinghai Lu, Yuning Mao, Xavier Martinet, Todor Mihaylov, Pushkar Mishra, Igor Molybog, Yixin Nie, Andrew Poulton, Jeremy Reizenstein, Rashi Rungta, Kalyan Saladi, Alan Schelten, Ruan Silva, Eric~Michael Smith, Ranjan Subramanian, Xiaoqing~Ellen Tan, Binh Tang, Ross Taylor, Adina Williams, Jian~Xiang Kuan, Puxin Xu, Zheng Yan, Iliyan Zarov, Yuchen Zhang, Angela Fan, Melanie Kambadur, Sharan Narang, Aur{\'{e}}lien Rodriguez, Robert Stojnic, Sergey Edunov,
  and Thomas Scialom. 2023.
\newblock \href {https://doi.org/10.48550/ARXIV.2307.09288} {L{L}a{MA} 2: Open foundation and fine-tuned chat models}.
\newblock \emph{CoRR}, abs/2307.09288.

\bibitem[{Vuli{\'c} and Mrk{\v{s}}i{\'c}(2018)}]{Vulic:2018naacl}
Ivan Vuli{\'c} and Nikola Mrk{\v{s}}i{\'c}. 2018.
\newblock \href {https://doi.org/10.18653/v1/N18-1103} {Specialising word vectors for lexical entailment}.
\newblock In \emph{Proceedings of the 2018 Conference of the North {A}merican Chapter of the Association for Computational Linguistics: Human Language Technologies, Volume 1 (Long Papers)}, pages 1134--1145, New Orleans, Louisiana. Association for Computational Linguistics.

\bibitem[{Vulić et~al.(2017)Vulić, Gerz, Kiela, Hill, and Korhonen}]{vulić2017hyperlex}
Ivan Vulić, Daniela Gerz, Douwe Kiela, Felix Hill, and Anna Korhonen. 2017.
\newblock \href {http://arxiv.org/abs/1608.02117} {Hyperlex: A large-scale evaluation of graded lexical entailment}.

\bibitem[{Wang et~al.(2022)Wang, Liu, Chen, Wang, and Zhao}]{wang-etal-2022-smash}
Yueqian Wang, Chang Liu, Kai Chen, Xi~Wang, and Dongyan Zhao. 2022.
\newblock \href {https://doi.org/10.18653/v1/2022.findings-emnlp.492} {{SMASH}: Improving {SMA}ll language models{'} few-{SH}ot ability with prompt-based distillation}.
\newblock In \emph{Findings of the Association for Computational Linguistics: EMNLP 2022}, pages 6608--6619, Abu Dhabi, United Arab Emirates. Association for Computational Linguistics.

\bibitem[{Ying et~al.(2018)Ying, He, Chen, Eksombatchai, Hamilton, and Leskovec}]{Ying_2018}
Rex Ying, Ruining He, Kaifeng Chen, Pong Eksombatchai, William~L. Hamilton, and Jure Leskovec. 2018.
\newblock \href {https://doi.org/10.1145/3219819.3219890} {Graph convolutional neural networks for web-scale recommender systems}.
\newblock In \emph{Proceedings of the 24th ACM SIGKDD International Conference on Knowledge Discovery; Data Mining}, KDD ’18. ACM.

\bibitem[{Zeng et~al.(2021)Zeng, Lin, Yu, Cleland{-}Huang, and Jiang}]{zeng2021enhancing}
Qingkai Zeng, Jinfeng Lin, Wenhao Yu, Jane Cleland{-}Huang, and Meng Jiang. 2021.
\newblock \href {https://arxiv.org/pdf/2106.02974} {Enhancing taxonomy completion with concept generation via fusing relational representations}.
\newblock In \emph{{KDD}}, pages 2104--2113. {ACM}.

\bibitem[{Zhang et~al.(2021)Zhang, Song, Zeng, Chen, Shen, Mao, and Li}]{zhang2021taxonomy}
Jieyu Zhang, Xiangchen Song, Ying Zeng, Jiaze Chen, Jiaming Shen, Yuning Mao, and Lei Li. 2021.
\newblock \href {http://www.arxiv.org/abs/2101.01896} {Taxonomy completion via triplet matching network}.
\newblock In \emph{Proceedings of the AAAI Conference on Artificial Intelligence}, volume~35, pages 4662--4670.

\end{thebibliography}
\bibliographystyle{acl_natbib}

\clearpage
\appendix
\onecolumn
\section{Using ChatGPT for Definition Generation and Automatic Error Analysis}\label{sec:chatgpt}

Here below are two different example prompts \ref{chatgpt_dataset} and \ref{chatgpt_dataset2} for ChatGPT for definition generation. The MAG PSY and MAG CS datasets for Taxonomy Enrichment, as well as ANT and HyperLex datasets for Lexical Entailment, do not possess definitions. Therefore, we developed several prompts specifically for the two types of datasets. For the hypernym prediction we want to have definitions for one input word, whereas for Lexical Entailment we expect to generate definitions for two words simultaneously, as they might be helpful for disambiguation. Table \ref{tab:def_chatgpt} represents statistics for the generated definitions for the datasets.

\lb{chatgpt_dataset}
{}
\begin{verbatim}
Write a definition for the word/phrase in one sentence.

Example:
Word: caddle
Definition: act as a caddie and carry clubs for a player

Word: eszopiclone 3 mg
Definition:
\end{verbatim}

\lb{chatgpt_dataset2}
{}
\begin{verbatim}
Write a definition for Word 1 and Word 2. Each definition should be 
in one sentence. If a word is ambiguous, use the other word to
disambiguate it.

Example:
Word 1: depression
Word 2: melancholy
Definition 1: a mental state characterized by a pessimistic sense of 
inadequacy and a despondent lack of activity
Definition 2: a constitutional tendency to be gloomy and depressed

Word 1: conflict
Word 2: disagreement
\end{verbatim}

\begin{table}[h]
    \centering
    \begin{tabular}{cccc}
    \toprule
       Dataset & Total & Generated with ChatGPT & From Wikidata  \\
    \midrule
      MAG PSY & 23,156 & 12,823 & 10,333  \\
      MAG CS & 29,484 & 5,714 & 23,770  \\
      ANT & 5,933 & 5,933 & -  \\
      HyperLex & 2,307 & 2,307 & -  \\
    \bottomrule
    \end{tabular}
    \caption{Statistics on definitions generated with ChatGPT for different tasks. }
    \label{tab:def_chatgpt}
\end{table}
\bigskip

Here is an example of the input to ChatGPT to automatically detect error types for hypernym prediction (Example \ref{chatgpt_input}) and the model output (Example \ref{chatgpt_output}). Furthermore, Example \ref{class_chatgpt} demonstrates an example prompt for automatic classification.

\lb{chatgpt_input}
{}
\begin{verbatim}
Here is the data in four columns separated with "\t":

topic: psychology	input word: social construction of schizophrenia
true: psychiatry, predictions: psychosis, social, psychological, 
and philosophical aspects, social psychology, psych, schizophrenia, 
antipsychotic, psychotic, antipsychotic

topic: psychology	input word: use heroin	true: methadone maintenance, 
predictions: heroin, opium tincture, heroin, opium, tincture, heroin, 
opium, and morphine, man, heroin, drug addiction

...

topic: psychology	input word: coordinated management of meaning	
true: social psychology, predictions: communication disorders,
language and literacy, and social studies, language and literacy, 
metaphor, metonymy, and politics, metonymy, metaphor

There is a topic, input word, correct hypernym (true), and model prediction 
(predicted). Your task is to compare true answers and model predictions 
and summarize what are the main error types of the model.
\end{verbatim}

%Then having collected pairs and definitions collected the prompt and target are the following:
\lb{chatgpt_output}{}
\begin{verbatim}
To summarize the main error types of the model's predictions compared 
to the true hypernyms:

1. Overly Broad Predictions: The model often generates predictions that 
encompass a broader concept than the true hypernym. 
    For example:
        "medication effects" was predicted with terms like "drug effects", 
        "drug interactions", and "medication", which are broader than 
        the true hypernym "clinical psychology" and "psychiatry."
        
2. Underly Broad Predictions: Conversely, some predictions are too narrow 
and fail to capture the broader concept represented by the true hypernym. 
    For example:
        "schizoid personality" was predicted with terms like "personality
        disorders" and "personality theory," which are narrower than the 
        true hypernym "psychosis."

3. Inclusion of Unrelated Terms: In some cases, the model includes terms 
in its predictions that are not directly related to the input word or
true hypernym. 
    For example:
        "pirisudanol" was predicted with terms like "dopamine receptor 
        d1" and "dopamine receptor d2", which are unrelated to the true 
        hypernym "psychiatry".
        
4. Repetition of Terms: The model occasionally repeats terms in its 
predictions, which might indicate redundancy or lack of diversity 
in its output. 
    For example:
        "pirisudanol" was predicted with repeated terms like 
        "piracetam" and "hydergine."
        
5. Inadequate Coverage of Concepts: Some input words and true 
hypernyms receive predictions that lack comprehensive coverage of
related concepts. 
    For example:
        "result category" was predicted with terms like "item 
        response theory" and "qualitative research," which might not 
        fully encapsulate the scope of the true hypernym "mathematics 
        education."
\end{verbatim}

%Then having collected pairs and definitions collected the prompt and target are the following:

\lb{class_chatgpt}
{}
\begin{verbatim}
You are a helpful linguist. Your task is to do error classification 
for a dataset for hypernym prediction in a general domain. Here are 
possible classes:

1. Overly Broad Predictions: The model often generates predictions 
encompassing a broader concept than the true hypernym. 
2. Underly Broad Predictions: Conversely, some predictions are too 
narrow and fail to capture the broader concept represented by the 
true hypernym.
3. Inaccurate Predictions: The model may predict words that are very 
semantically close to the true hypernym, but struggles with fitting 
into the exact wording 
4. Conceptual Ambiguity: The model may struggle with ambiguous 
(polysemantic/multivalued) input words or concepts, leading to 
incorrect predictions.
5. Incorrect definitions: The model gets confused with the incorrect/
inaccurate definition retrieved from external sources

You will be given an input word/phrase, true hypernym, and
candidate hypernyms. Please, return a Python dict of error classes 
{1: 1, 2: 5, 3: 1, ..., 100:3}) for all instances below:

id: 1, input word: parathyroid_hormone, true hypernym: hormone, 
predicted: hormonal agent, hormon, hematopoietic growth factor,
growth factor of the blood, growth regulator, growth substance, growth
...
id: 100, input word: proofreader, true hypernym: printer, predicted: 
reader, audience, audience member, spectator, viewer, listener, 
listener-in, hearer, recipient, witness, watcher, observer
\end{verbatim}

\bigskip

Here is the prompt Example \ref{class_chatgpt2} for ChatGPT in order to automatically evaluate \text{TaxoLLaMA} results, as manual analysis has shown that the gold true answers from MAG PSY and MAG CS datasets might not be of a good quality either. Therefore, ChatGPT was required to choose between the true answer from the dataset and the predicted candidate.

\lb{class_chatgpt2}
{}
\begin{verbatim}
    Here are the words in the psychological domain. Your task is to
    choose hypernym which is more relevant given two options. 
    Answer 1 / 2 / both / none

Example:
social construction of schizophrenia
option 1: psychosis 
option 2: psychiatry
Answer: 2

abdominal air sac
option 1: air sac
option 2: trachea
Answer:
\end{verbatim}

\begin{table}[h]
    \centering
    \begin{tabular}{p{0.35\linewidth} p{0.6\linewidth}}
    \toprule
      Error Type   & Descripton \\
      \midrule
       Overly Broad Predictions  & The model often generates predictions that encompass a broader concept than the true hypernym.\\\addlinespace 
       Underly Broad Predictions & Some predictions are too narrow and fail to capture the broader concept represented by the true hypernym \\
    \addlinespace 
    Inclusion of Unrelated Terms & In some cases, the model includes terms in its predictions that are not directly related to the input word or true hypernym. \\ \addlinespace
    Repetition of Terms & The model occasionally repeats terms in its predictions, which might indicate redundancy or lack of diversity in its output. \\
\addlinespace
Inadequate Coverage of Concepts & Some input words and true 
hypernyms receive predictions that lack comprehensive coverage of
related concepts \\ \addlinespace
Semantic Shift & The model might exhibit errors related to semantic shift, where the predicted terms are semantically related to the input word but do not accurately reflect the intended meaning or context. \\ \addlinespace
Conceptual Ambiguity &  The model may struggle with ambiguous input words or concepts, leading to predictions that lack clarity or specificity.  \\
\addlinespace Domain-Specific Knowledge & Errors may arise due to a lack of domain-specific knowledge or understanding of specialized terminology. \\ \addlinespace
Cultural or Contextual Bias & The model's predictions may be influenced by cultural or contextual biases inherent in the training data. This could lead to inaccuracies, especially when dealing with topics or concepts that vary across cultures or contexts.\\ \addlinespace
Incomplete Understanding of Relationships & The model may struggle to understand complex relationships between concepts, leading to inaccurate predictions.\\
\addlinespace
Word Sense Disambiguation & Errors may occur due to difficulties in disambiguating between different senses of a word. \\
\addlinespace Knowledge Gap &  The model's predictions may reflect gaps in its knowledge or understanding of certain concepts, resulting in inaccurate or incomplete responses. \\
    \bottomrule
    \end{tabular}
    \caption{ 12 Error types made by \text{TaxoLLaMA} for hypernym prediction detected by ChatGPT.}
    \label{tab:chatgpt}
\end{table}

\section{Error Type Analysis}\label{sec:erorr_type_analysis}

This section represents the error types distribution across different datasets for hypernym prediction: Hypernym Discovery and
Taxonomy Enrichment in Table \ref{tab:errors_distribution}. Moreover, in Table \ref{tab:examples} we provide an example for each error type that was classified by ChatGPT.

\begin{table}[ht!]
\centering
\resizebox{0.98\textwidth}{!}{
\begin{tabular}{lccccc|cccc}
\toprule
 & 1A: English & 2A: Medical & 2B: Music & 1B: Italian & 1C: Spanish & MAG-CS & MAG-PSY & Noun & Verb \\
\midrule
Error 1 & 72.49\% & 93.75\% & 100.0\% & 54.69\% & 49.08\% & 66.48\% & 85.43\% & 81.45\% & 73.39\% \\
Error 2 & 2.61\% & 0.00\% & 0.0\% & 10.03\% & 10.62\% & 5.40\% & 1.40\% & 4.10\% & 2.58\% \\
Error 3 & 9.44\% & 1.25\% & 0.0\% & 27.83\% & 12.09\% & 13.21\% & 11.20\% & 6.05\% & 21.03\% \\ 
Error 4 & 15.46\% & 5.00\% & 0.0\% & 7.44\% & 18.21\% & 10.94\% & 0.70\% & 6.64\% & 0.43\% \\
Error 5 & 0.00\% & 0.00\% & 0.00\% & 0.00\% & 0.00\% & 3.98\% & 1.26\% & 1.76\% & 2.58\% \\
\bottomrule
\end{tabular}
}
\caption{Errors Type Distribution across subset datasets for hypernym prediction: Hypernym Discovery and Taxonomy Enrichment.}
\label{tab:errors_distribution}
\end{table}

\begin{table}[h!]
    \centering
    \begin{tabular}{p{0.2\linewidth}p{0.2\linewidth}p{0.2\linewidth}p{0.3\linewidth}}
    \toprule
      Error Type   & Input & True & Predicted \\
      \midrule
       Overly Broad Predictions  &  weed & vascular plant & plant, plant life, vegetable, vegetation, vascular plant, phanerogam, green plant, green thing, plant of the class, plant of the division \\ \addlinespace 
       Underly Broad Predictions &  audio & auditory communication & signal, transmission, transmission signal, communication, communication signal, message, message signal, information, information signal, data, data signal, notification, notification signal, signaling
 \\
    \addlinespace 
    Inaccurate Predictions & unconventionality & unorthodoxy & unorthodoxiness, unorthodxy, unconformity, heterodoxy, dissidence, disobedience, disloyalty, dis, disaffection, dissent, disagreement \\ \addlinespace
    Conceptual Ambiguity & giant red paintbrush & indian paintbrush & paintbrushrush, paint brush, larkspur, delphinium, liliaceous plant, liliid monocot, lilioid \\
\addlinespace
Incorrect definitions & advance directive status
& documentation
& advance directive, directive, decision support system, clinical decision support system, health information technology \\
\multicolumn{4}{l}{\multirow{2}{*}{\begin{tabular}[c]{@{}l@{}}\textit{Definition: A do-not-resuscitate order (DNR), also known as Do Not Attempt Resuscitation (DNAR),}\\ \textit{Do Not Attempt Cardiopulmonary Resuscitation (DNACPR)}\end{tabular}}} \\
     \\
     \bottomrule
    \end{tabular}
    \caption{Examples for each  Error type made by \text{TaxoLLaMA} for hypernym prediction detected by ChatGPT.}
    \label{tab:examples}
\end{table}

\section{Distribution Visualization for Taxonomy Enrichment}
\label{sec:appendix_tsne}

\begin{figure*}[ht!]
    \centering
    \includegraphics[width=\textwidth]{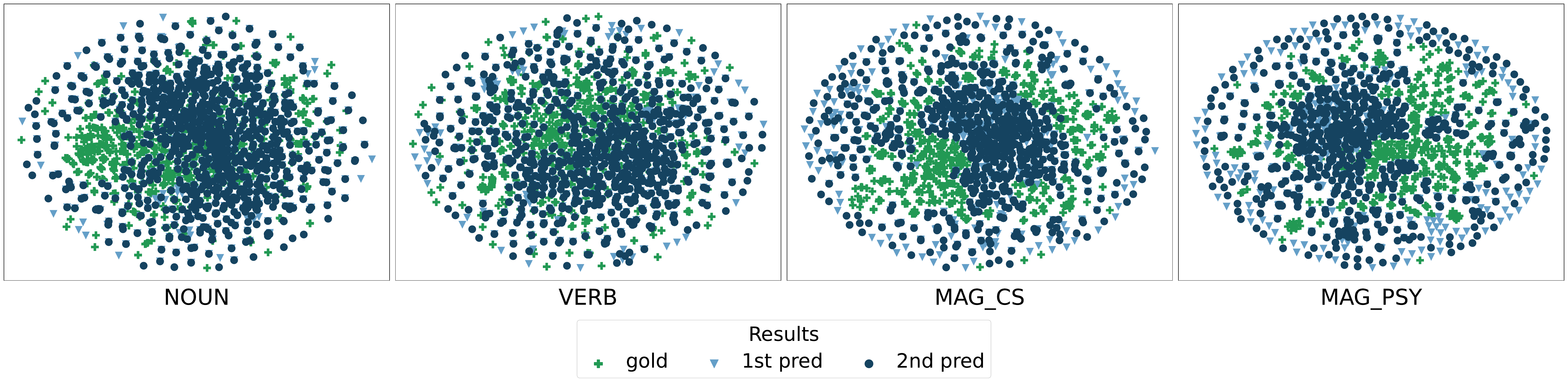}
    \caption{t-SNE plot of distributions of ground truth nodes and predicted nodes for taxonomy enrichment tasks. Each point represents a node, embedded with SentenceBert. Color represents ground truth or model predictions (we ran 2 predictions with different seeds)}
    \label{fig:enrich_analysis}
\end{figure*}

In this section, we delve into the distribution patterns of ground truth and model predictions within the embedding space of the SentenceBert model \cite{reimers2019sentencebert}. To achieve this, we initiated two separate model runs, each with a distinct seed, aiming to capture the model's variability. Subsequently, we extracted the predicted candidates and the ground truth hypernyms, mapping them into the embedding space provided by SentenceBert. To facilitate a clearer visual analysis, we condensed the embedding dimensions to 50 using Principal Component Analysis and then applied t-SNE to project these dimensions onto two principal components for visualization.

The findings, illustrated in Figure \ref{fig:enrich_analysis}, reveal a distinct pattern between WordNet and the MAG subsets (MAG\_CS and MAG\_PSY). WordNet displays a notable overlap between the gold standard and predictions, despite a few outliers that are presumably lower-ranked candidates. Conversely, the MAG subsets exhibit different behavior, forming two slightly overlapping clusters in the embedding space, suggesting a divergence between predictions and ground truths. Additionally, these subsets contain more outliers, indicating instances where the model may have completely missed the accurate hypernym sense. It's important to consider, however, that the SentenceBert model's representations could contribute to these discrepancies, especially for concepts that are not well-represented in its training data.

\section{Hyperlex Correlation Analysis}
\label{sec:appendix_corr}

\begin{figure}[h]
\centering
\includegraphics[width=0.7\linewidth]{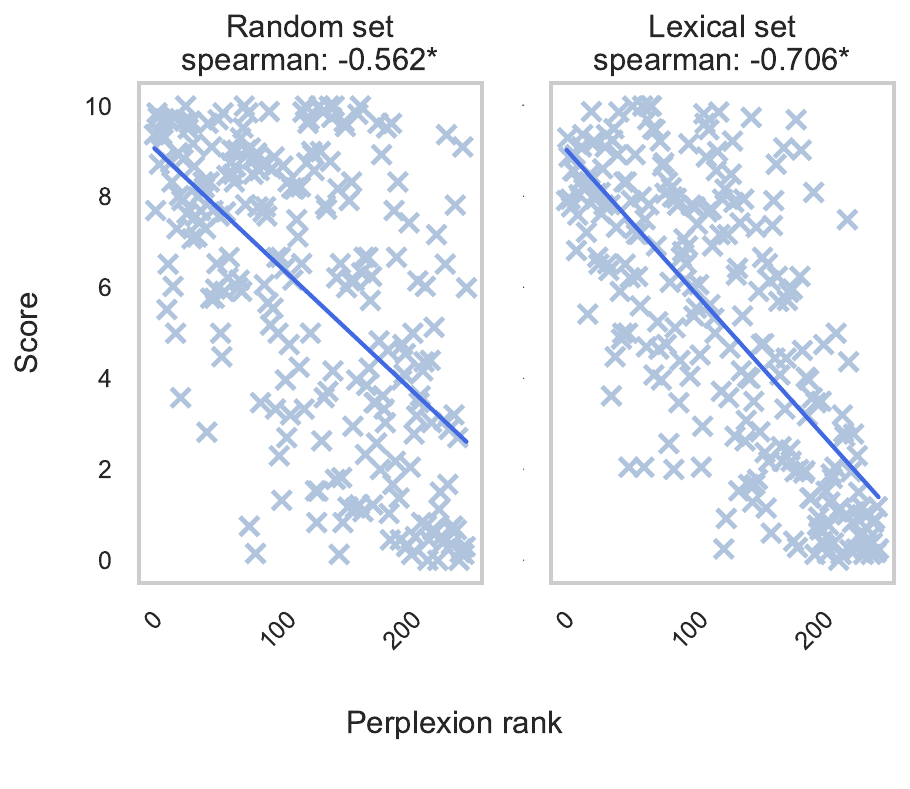}
\caption{Correlation plot of the perplexion ranks with the annotator's score on Hyperlex test sets. The line over the dots is a trend found with linear regression. * shows that correlation has a p-value lower than $1e^{-4}$.}
\label{fig:mpr}
\end{figure}

We also examine correlations using traditional methods for both test sets (refer to Figure~\ref{fig:mpr}). By overlaying the linear regression trend on the observed data points, a distinct trend emerges. However, this trend is notably impacted by outliers, particularly within the Random set. This observation aligns with findings from taxonomy construction, highlighting the model's challenges in accurately handling middle nodes or pairs exhibiting moderate entailment strength.

When analyzing gold scores ranging from 2 to 8, the Random set displays a lack of discernible trend, underscoring the model's inconsistency in this area. The Lexical set shows a slightly better trend in that area. However, with both sets pairs characterized by strong entailment or minimal entailment are more accurately categorized. This distinction crucially enhances the overall correlation, which leads to a promising correlation score.

\section{Hypernym motivation} \label{sec:appendix_hypernym}

Inspired by recent advancements in semantic analysis, particularly the work \citep{nikishina-etal-2023-predicting} on hyponym prediction, our study shifts focus towards hypernym prediction for several compelling reasons. First, predicting hypernyms is crucial for tasks such as taxonomy enrichment and hypernym discovery. Second, the formulation of a loss component for hypernym prediction is more straightforward, as most entities typically have a single correct hypernym, unlike hyponyms, where multiple valid options exist. This necessitates either adjustments to the loss function or extensive dataset collection and analysis.

Furthermore, our experimentation with various prompts revealed that the most effective format is detailed in the main section. Alternate prompts, which adopted a more narrative style (e.g., ``Given a hyponym 'tiger', the hypernym for it is"), led to the model generating paragraphs instead of concise hypernym lists. Adjustments to the system prompt failed to rectify this. Notably, appending a comma to the end of the target sequence remarkably improved the model output, encouraging it to list hypernyms instead of producing narrative text.

In addressing the disambiguation challenge, we experimented with incorporating definitions or technical identifiers from WordNet into the prompts. Definitions proved more effective, likely owing to the model pre-training on textual data. Attempts to generate hypernyms with specific WordNet codes resulted in the model appending the same numerical identifier to each hypernym which also resulted in lower scores.

\section{Hyperparameter motivation} 
\label{sec:appendix_hyperparams}

Our analysis revealed the model acute sensitivity to the learning rate and scheduler settings. The feasibility of employing a high learning rate in the primary study was contingent upon the use of the LORA adapter, which modulates weights without significant alterations. However, during full model fine-tuning, we faced instabilities, manifesting as either overfitting or underfitting—highlighting the necessity for further technical exploration into optimal hyperparameter configurations. Additionally, the implementation of 4-bit quantization requires careful learning rate selection, as this process notably compresses the weight distribution, demanding strategies to effectively recover the model knowledge thereafter.

In the fine-tuning process, we deliberately chose a smaller batch size to better accommodate the model to datasets, which are often limited in sample size. Contrary to our expectations, increasing the learning rate and batch size did not yield improved performance; this outcome can primarily be attributed to the reduced number of steps the model takes toward adapting to the specific domain. This strategy, however, did not apply to WordNet pre-training, where we observed differing trends.

Apart from certain instruction tuning methodologies, our approach does not involve calculating loss including on the instruction itself. Instead, loss calculation is confined solely to the target tokens.

The experiments utilized Nvidia A100 or Quadro RTX 8000 GPUs. Pre-training for \text{TaxoLLaMA} and \text{TaxoLLaMA-bench} spanned 6 GPU hours, while \text{TaxoLLaMA-verb} required less than 1 hour. Fine-tuning for MAG subsets took 5 GPU hours, attributed to the lengthy definitions. Fine-tuning for other datasets was completed in under an hour.

\end{document}